\documentclass{amsart}
\usepackage{graphicx} % Required for inserting images
\usepackage{amsfonts, amsmath, amssymb, amsthm}
\usepackage{hyperref}
\usepackage{subcaption}
\usepackage{url}
\usepackage{bbold}
\makeatletter
\def\amsbb{\use@mathgroup \M@U \symAMSb}
\makeatother

\theoremstyle{definition}
\newtheorem{definition}{Definition}[section]

\theoremstyle{remark}
\newtheorem*{remark}{Remark}
\newtheorem{approach}{Approach}

\newcommand{\E}{\amsbb{E}}
\newcommand{\Ttrans}{T_{\mathrm{transformed}}}
\newcommand{\Tmid}{T_{\mathrm{midpoint}}}
\title{Causal Inference in Energy Demand Prediction}

\author[Ma]{Chutian Ma}
\email{c.ma@causify.ai}
\author[Pomazkin]{Grigorii Pomazkin}
\email{g.pomazkin@causify.ai}
\author[Saggese]{Giancinto Paolo (GP) Saggese}
\email{gp@causify.ai}
\author[Smith]{Paul Smith}
\email{paul@causify.ai}

\begin{document}

\begin{abstract}
    Energy demand prediction is critical for grid operators, industrial energy
    consumers, and service providers. Energy demand is influenced by multiple
    factors, including weather conditions (e.g. temperature, humidity, wind
    speed, solar radiation), and calendar information (e.g. hour of day and
    month of year), which further affect daily work and life schedules. These
    factors are causally interdependent, making the problem more complex than
    simple correlation-based learning techniques satisfactorily allow for. We
    propose a structural causal model that explains the causal relationship
    between these variables. A full analysis is performed to validate our causal
    beliefs, also revealing important insights consistent with prior studies.
    For example, our causal model reveals that energy demand responds to
    temperature fluctuations with season-dependent sensitivity. Additionally, we
    find that energy demand exhibits lower variance in winter due to the
    decoupling effect between temperature changes and daily activity patterns.
    We then build a Bayesian model, which takes advantage of the causal insights
    we learned as prior knowledge. The model is trained and tested on unseen
    data and yields state-of-the-art performance in the form of a 3.84\% MAPE on
    the test set. The model also demonstrates strong robustness, as the
    cross-validation across two years of data yields an average MAPE of 3.88\%.
\end{abstract}

\maketitle
\tableofcontents

\section{Introduction}
While machine learning (ML) has achieved enormous success in recent decades,
most modern machine learning systems operate purely on statistical correlation,
without any understanding of causation. They excel at detecting patterns in
training data, but cannot distinguish between relationships that merely happen
to occur together and those that represent genuine cause-and-effect mechanisms.
As Bernhard pointed out in \cite{2022Schoelkopf}, this distinction represents
one of the most profound gaps between current AI capabilities and human
intelligence and it hinders the ability of AI systems to generalize what they
have learned to new problems. 

Due to the these limitations of correlation-based AI, the machine learning
community has shown increasing interest in incorporating causal inference in
machine learning \cite{kaddour2022causal, yao2021survey, jiao2024causal,
brand2023recent}, and particularly for time series prediction
\cite{moraffah2021causal, runge2023causal}. 

In terms of ``hindering the generalization ability", the presence of confounder
bias stands as one of the most common challenges. In this paper, we define a
confounder in the same way as in Pearl's work on causality (e.g., \cite{pearl2009causality}).
\begin{definition}[Confounder]\label{def:confounder}
    Suppose the random variables $X, Y, Z$ are causally connected by the
    following relation $X \leftarrow Z \rightarrow Y$ ($Z$ affects $X$ and $Y$
    causally). Then we say that $Z$ confounds $X$ and $Y$, or that $Z$ is a
    confounder of the other two variables.
\end{definition}

The existence of confounders is often problematic because they create spurious
associations between predictors and outcomes. Models learn these non-causal
correlations rather than genuine causal effects. This hinders generalization:
when the distribution of predictors shifts in deployment environments or under
interventions, the spurious patterns learned from confounded training data fail
to hold, leading to poor predictive performance in new settings. This is a major
limitation of purely statistical approaches operating at the associational
level, as Pearl emphasizes in his discussion of the causal hierarchy
\cite{2018Pearl}.

In this paper, we study an energy demand prediction problem in which we model
the system load using ML approaches. We present a full analysis of the
interdependency structure of the various predictors (including weather
conditions and calendar information) and energy demand. At the same time, we
list common mistakes and consequences originating from confounder bias and
misspecified causal structure. A Bayesian causal model is later built based on
the causal insights, trained on real-world energy demand data and tested on
unseen data. The model is able to produce state-of-the-art predictions, with an
average MAPE of 3.88\% generated from a $K$-fold cross validation test. In
addition, the model is able to explain the variability found in the data, such
as the seasonal-dependent variance (heteroscedasticity) and temperature
sensitivity.

This paper is structured as follows. Section \ref{sec:causal vs non causal}
presents a full analysis of the interdependency between calendar, weather and
energy demand variables. We show that ignoring causal structure can cause a
trained model to show confounder bias, severely jeopardizing its generalization
to unseen data. Section \ref{sec: bayesian model} develops a Bayesian model for
causal inference and derives insights into the causal dependencies of energy
demand. Sections \ref{sec:conclusion} and \ref{sec:future directions} summarize
our findings and propose future research directions to improve the accuracy of
the model. Appendix \ref{sec: appendix} provides a formal proof using the
backdoor criterion to verify that the model used in Section \ref{sec: bayesian
model} is causally justified in the sense that no confounder biases are present
in the resulting estimates.

\section{Analysis of Causal Structure}\label{sec:causal vs non causal}
%------Generic introduction of what influence the energy demand ----
%------and how causality manifests----------------------------------

It is well known that calendar variables and weather conditions are among the
most useful predictors of electricity consumption. For example, during hot
summer days, air conditioning usage drives demand upward, while cold winter days
increase heating loads. Similarly, electricity usage follows distinct patterns
throughout the day, regardless of prevailing weather conditions, with commercial
loads peaking during business hours and residential consumption rising in the
evenings and weekends. Understanding these causal relationships--rather than
merely observing their correlations--is essential for building robust prediction
models that generalize across seasons, weather conditions, and changing usage
patterns. Failure to recognize the underlying causal structure may lead to
incorrect attributions of individual variables to the total effect, hindering
model generalization in cases where the underlying distributions shift. In this
section, we propose a causal structure that explains the interdependency between
weather, calendar variables, and energy demand. We show that missing certain
variables in feature selection not only causes models to lose the predictive
power that these variables provide, but may also prevent models from correctly
learning the effects of the variables that were selected.

\subsection{Dataset Description}
We obtained publicly available electricity load data from September 2023 to
August 2025 from the Upper Great Plains East (WAUE) balancing authority, which
belongs to the Southwest Power Pool \cite{spp2025hourlyload}. To isolate model
performance from weather forecast uncertainty, we evaluate models using actual
observed weather data.

The historical weather we obtained from the Open-Meteo Historical Weather API
\cite{openmeteoapi}, which provides ERA5 reanalysis data \cite{hersbach2020era5}
at hourly resolution. The weather variables include temperature (2 meters from
the ground), wind speed (10 meters from the ground), relative humidity (2 meters
from the ground) and solar radiation. For simplicity, weather from the location
LATITUDE = 44.6321, LONGITUDE = -100.2753 is chosen as representative of the
whole Upper Great Plains East region.

%------------------------Introduce our model -----------------------
\subsection{Structural Causal Model}
\begin{figure}[htbp]
    \centering
    \includegraphics[width=0.8\linewidth]{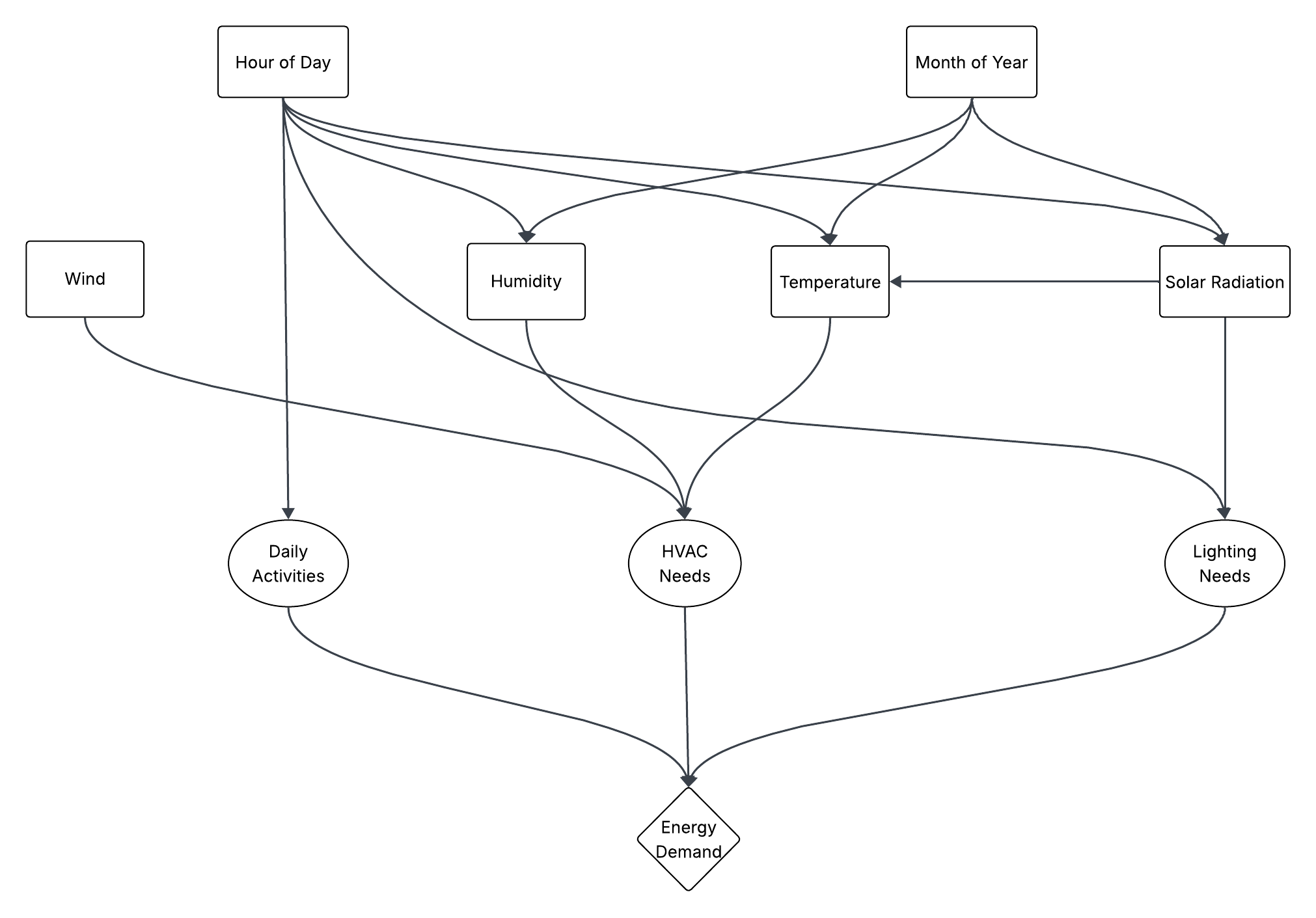}
    \caption{Structural Causal Model Driven by Calendar and Weather}
    \label{fig:updated_energy_dag}
\end{figure}

% 1. Review what is structural causal model 
Our causal treatment is largely based on Pearl's causality framework. The causal
mechanism between variables is represented as a directed acyclic graph (DAG),
where each node represents a variable and a functional relationship is assigned
to each edge, which characterizes the influence of a cause on its effect. 
% 2. Propose our DAG
We consider two major classes of variables: calendar variables and weather
variables:
\begin{enumerate}
    \item Calendar variables include hour of day and month of year.
    \item Weather variables include temperature, humidity, wind speed and solar radiation.
\end{enumerate}  
While these variables jointly drive energy demand, their contribution to the
total demand can be divided into three major categories, which are
non-intersecting: (1) energy consumption due to people's routine activities;
(2) heating, ventilation, and air conditioning (HVAC) needs; and (3) lighting needs:
\begin{enumerate}
    \item Routine activity needs are considered to be the base level of energy
    consumption during people's activities, e.g. industrial activities,
    entertainment, etc. ``Base level" means that we do not factor in additional
    energy costs due to heating or cooling needs in the case of extreme weather
    or additional lighting needs during evening times.
    \item HVAC needs represent the additional energy consumption due to heating,
    ventilation and air conditioning.
    \item Lighting needs represent the additional energy consumption due to
    insufficient light (e.g., due to seasonal and hour-of-day effects).
\end{enumerate}
We propose the DAG found in Figure \ref{fig:updated_energy_dag} to describe the
causal mechanism among these variables.
The nodes in rectangular blocks in Figure \ref{fig:updated_energy_dag} are the
observed variables. The nodes in oval blocks represent three major categories of
energy demand. Each arrow represents a cause-and-effect relationship. For
example, HVAC needs are jointly determined by temperature, humidity and wind,
while these weather variables themselves are affected by the calendar variables
hour and month.

\subsection{Humidity}\label{subsec:humidity}
\begin{figure}[htbp]
    \centering
    \includegraphics[width=0.8\linewidth]{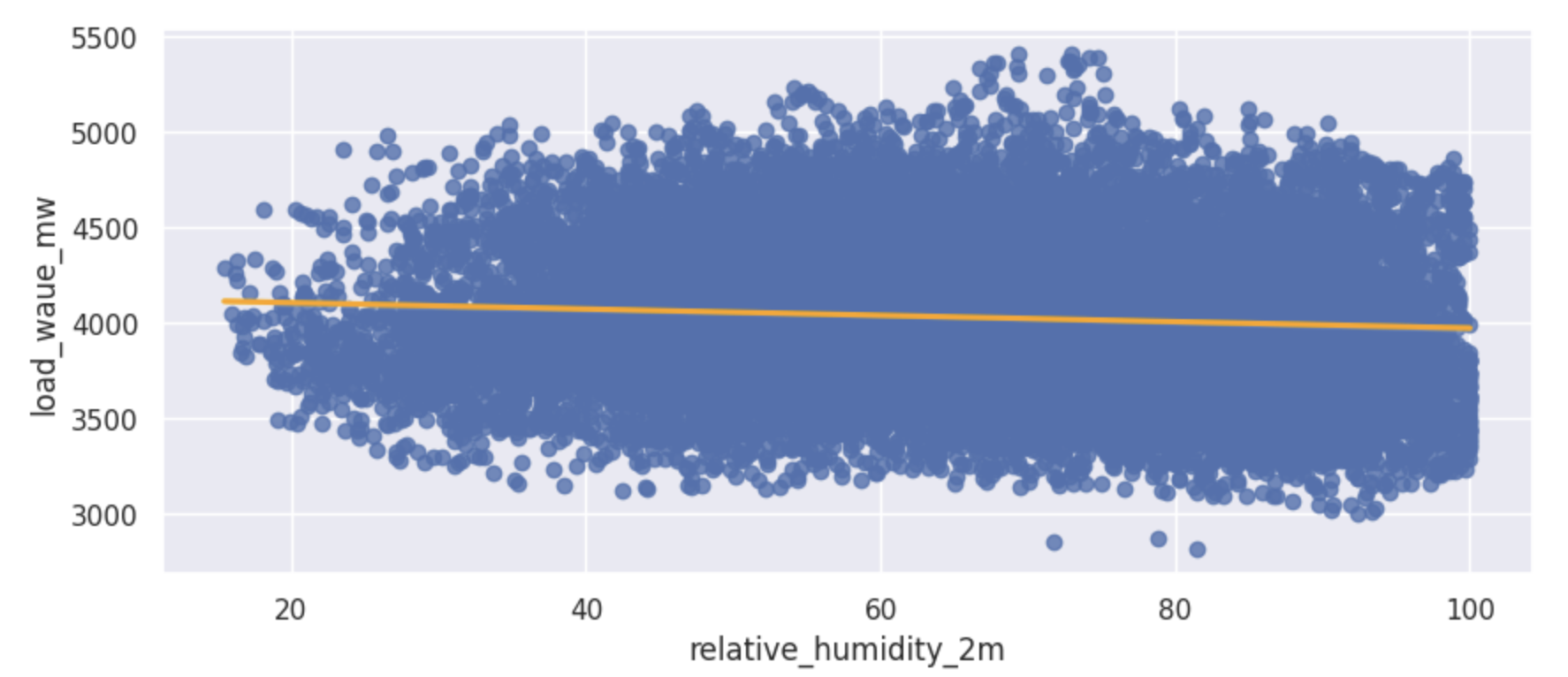}
    \caption{Humidity vs. Energy Demand}
    \label{fig:humidity_vs_energy_demand}
\end{figure}

\begin{figure}[htbp]
    \centering
    \includegraphics[width=0.8\linewidth]{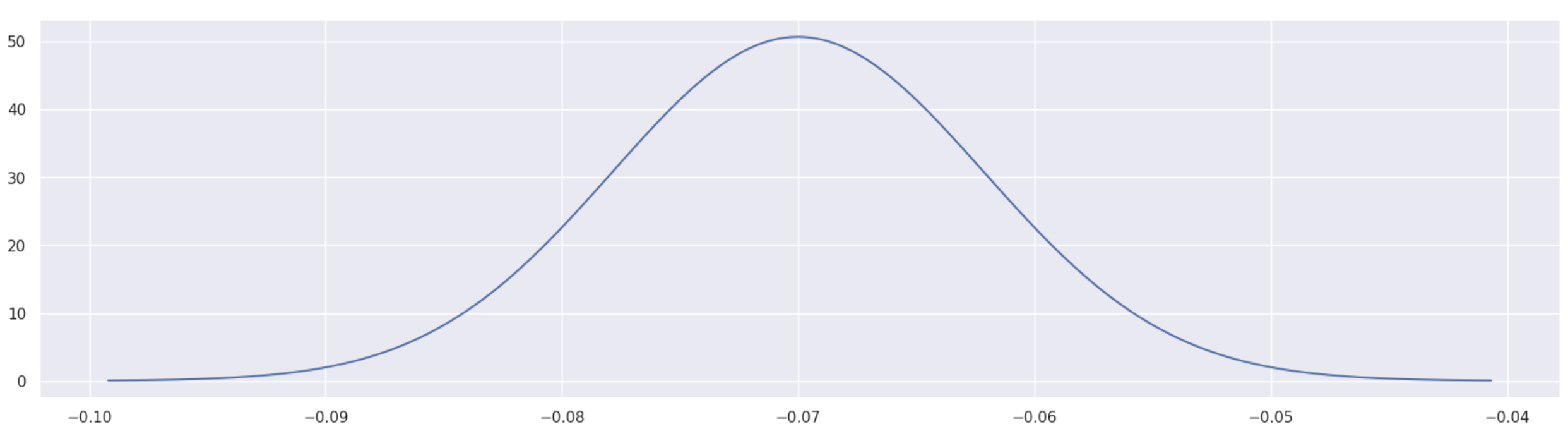}
    \caption{The Distribution of Correlation}
    \label{fig:humidity_corr_distribution}
\end{figure}

In traditional modeling practice, one would only care about the functional
relationship between the predictors and the target. Learning relations between
predictors is not a typical explicit goal of traditional modeling. However,
failing to capture causal effects between predictors, as traditional machine
learning techniques fail to do, can lead to erroneous inferences. For example,
if one examines the relation between humidity data and energy demand in the WAUE
dataset, one will initially notice a negative correlation, as revealed by Figure
\ref{fig:humidity_vs_energy_demand}.
In fact, we have
\begin{equation}\label{eq:incorrect humidity vs energy corr}
    \text{corr}(\text{humidity}, \text{energy}) = -0.07.
\end{equation}
To quantify uncertainty in the estimated correlation \eqref{eq:incorrect
humidity vs energy corr}, we compute (see Figure
\ref{fig:humidity_corr_distribution}) the approximate confidence density using
Fisher's z-transform, which provides an excellent approximation to the exact
posterior in our regime (see equation (2.11) and discussion in
\cite{taraldsen2023confidence}).

\begin{figure}[htbp]
    \begin{subfigure}[b]{\textwidth}
        \centering
        \includegraphics[scale=0.3]{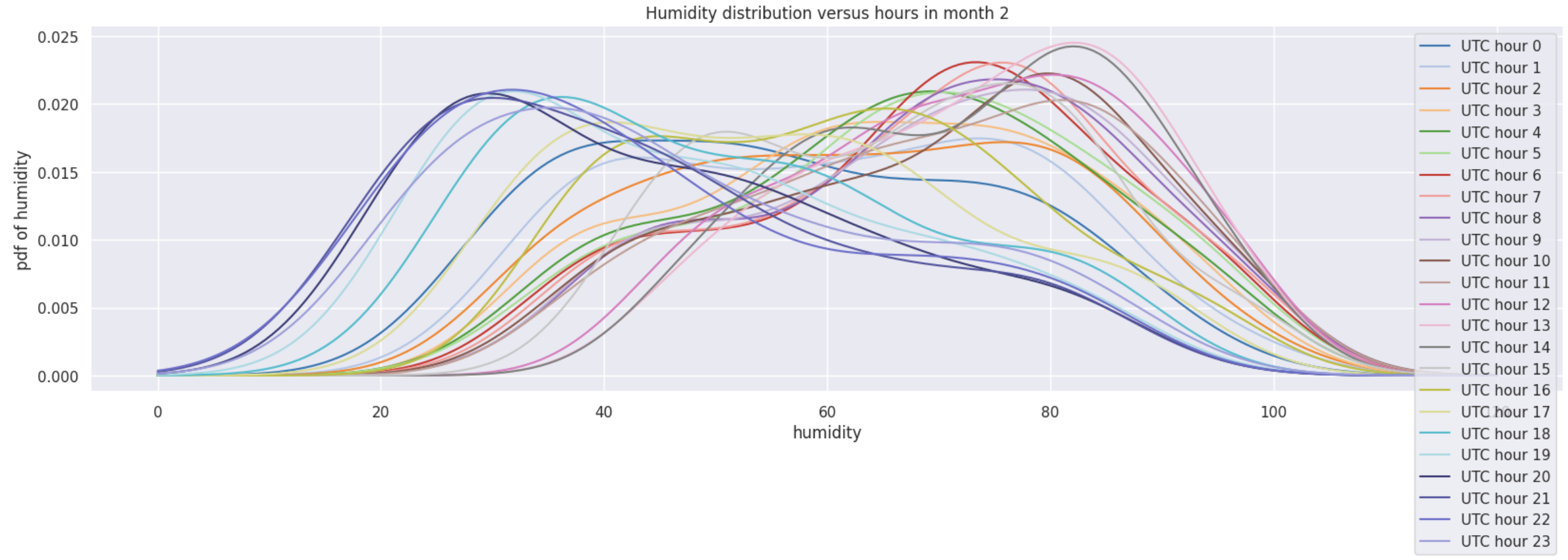}
        \caption{January}
        \label{fig:humidity_kde_Jan}
    \end{subfigure}
    \vspace{0.5cm}
    \begin{subfigure}[b]{\textwidth}
        \centering
        \includegraphics[scale=0.3]{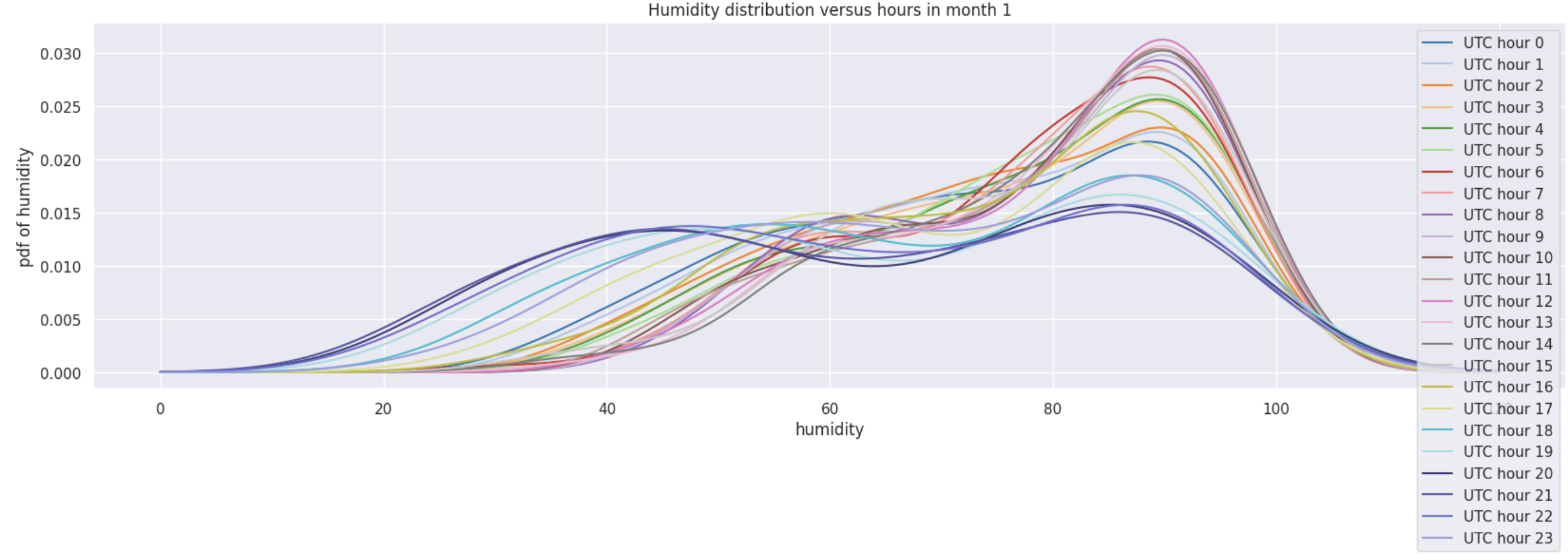}
        \caption{February}
        \label{fig:humidity_kde_Feb}
    \end{subfigure}
    \caption{The Distribution of Humidity Against Month and Hour}
    \label{fig:humidity_distribution_vs_hour}
\end{figure}

However, this contradicts domain knowledge, which suggests that higher humidity
increases ventilation and cooling needs. This contradiction occurs because we
ignored the variable hour of day, which is a confounder of humidity and energy
demand, during feature selection. We can verify that humidity is causally
affected by hour of the day by examining its conditional distribution, as shown
in figure \ref{fig:humidity_distribution_vs_hour}. Each curve corresponds to the
density function of humidity sampled from a specific hour of the day. 

\begin{figure}[htbp]
    \centering
    \includegraphics[width=0.4\linewidth]{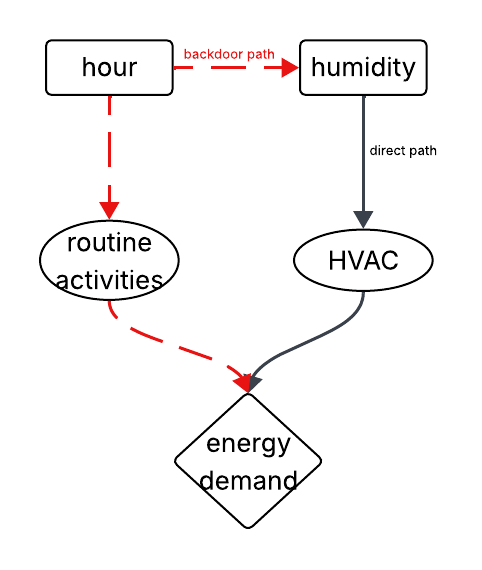}
    \caption{Direct Path and Backdoor Path}
    \label{fig:humidity_partial_DAG}
\end{figure}

Note that humidity reaches its peak in the morning around 6 a.m. Despite the
fact that high humidity tends to increase ventilation and cooling needs, 6 a.m.
is an off-peak hour in terms of energy consumption attributable to people's
routine activities. As a result, conditioning on humidity directly increases the
HVAC demand, while at the same time has the ``unexpected" effect of shifting the
time toward an off-peak hour, which indirectly lowers the energy consumption
cost. In the end, these two effects sum up to an overall negative correlation as
we have seen in Figure \ref{fig:humidity_vs_energy_demand}. The intuition can be
visualized in Figure \ref{fig:humidity_partial_DAG} where the black solid path
``humidity$\rightarrow$HVAC$\rightarrow$energy" represents the causal effect of
humidity on the energy demand and the red dotted path
``humidity$\leftarrow$hour$\rightarrow$routine activities$\rightarrow$energy"
represents the implicit effect caused by conditioning on humidity. 

\begin{figure}[htbp]
    \centering
    \begin{subfigure}[b]{0.48\textwidth}
        \centering
        \includegraphics[width=\textwidth]{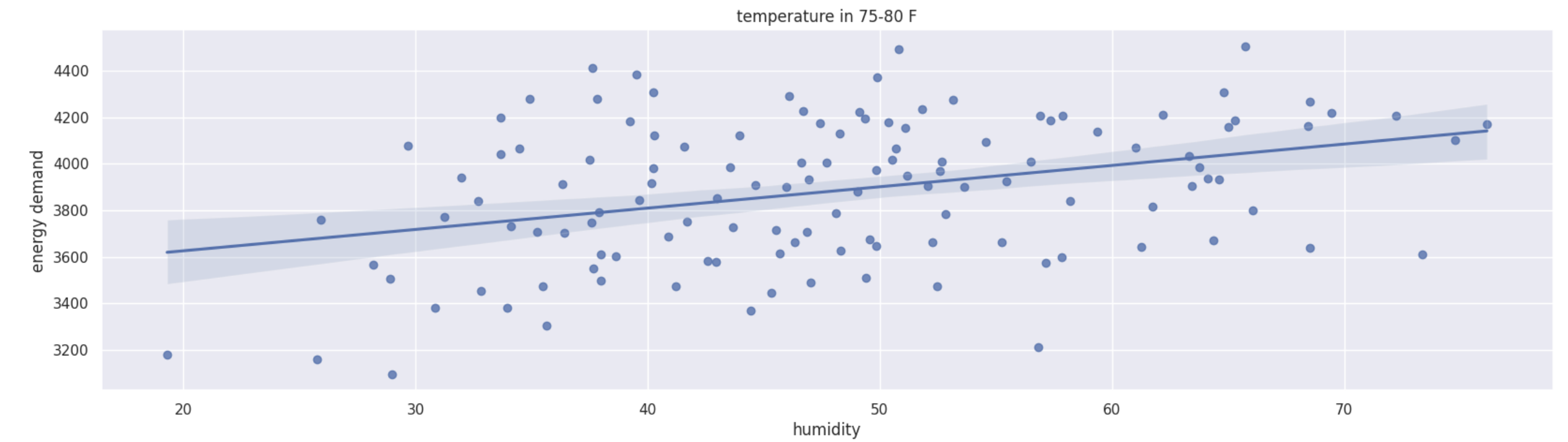}
        \vspace{0.5cm}  % Adjust vertical spacing between images
        \includegraphics[width=\textwidth]{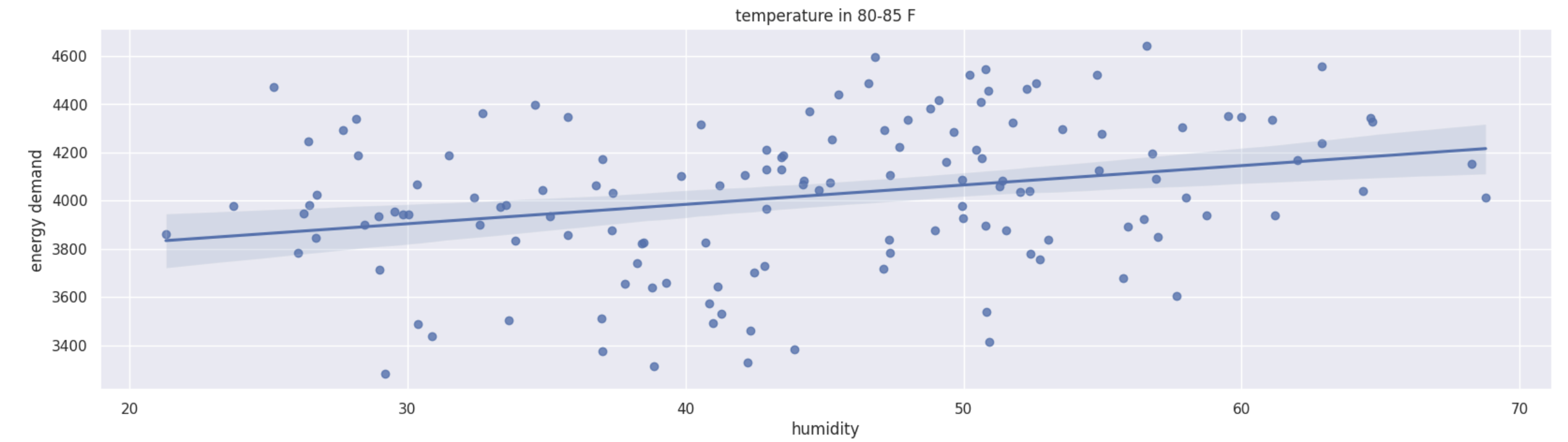}
        \vspace{0.5cm}  % Adjust vertical spacing between images
        \includegraphics[width=\textwidth]{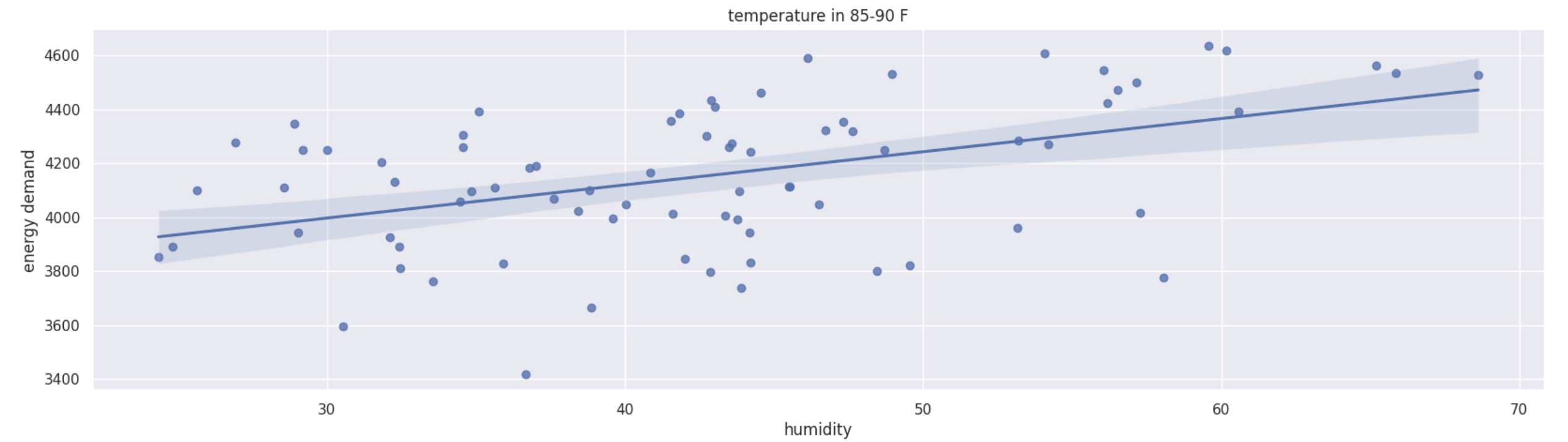}
        \caption{High Temperature Scenario}
    \end{subfigure}
    \hfill
    \begin{subfigure}[b]{0.48\textwidth}
        \centering
        \includegraphics[width=\textwidth]{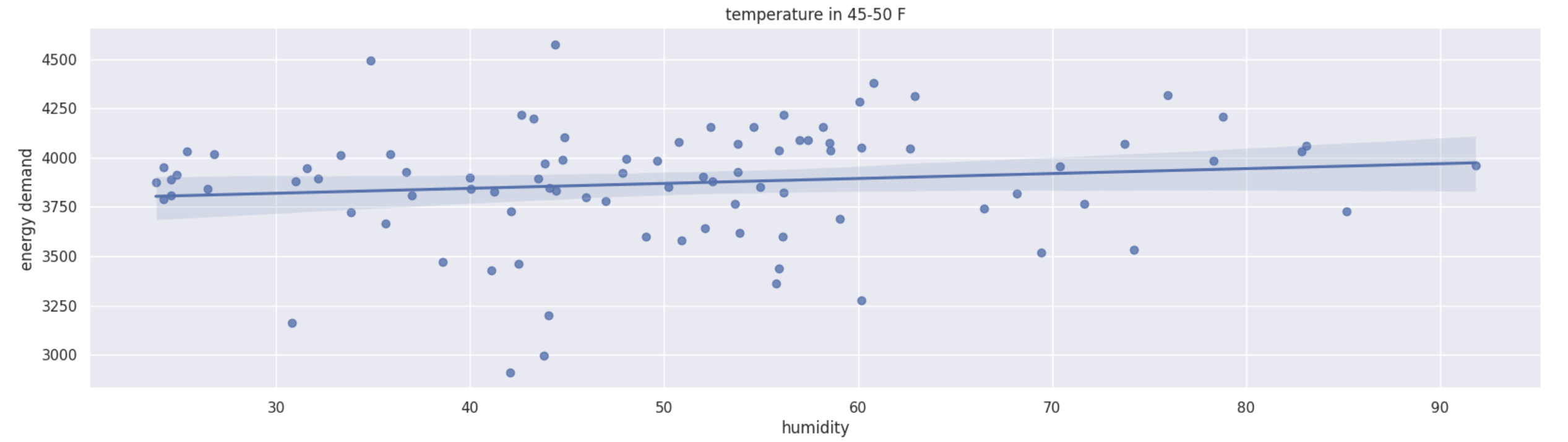}
        \vspace{0.5cm}  % Adjust vertical spacing between images
        \includegraphics[width=\textwidth]{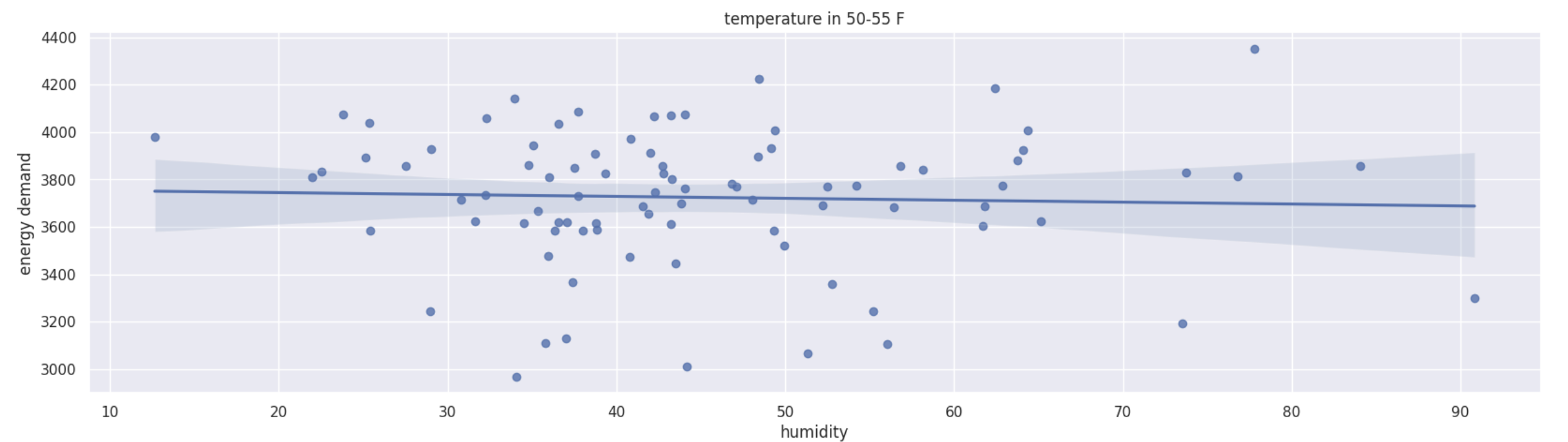}
        \vspace{0.5cm}  % Adjust vertical spacing between images
        \includegraphics[width=\textwidth]{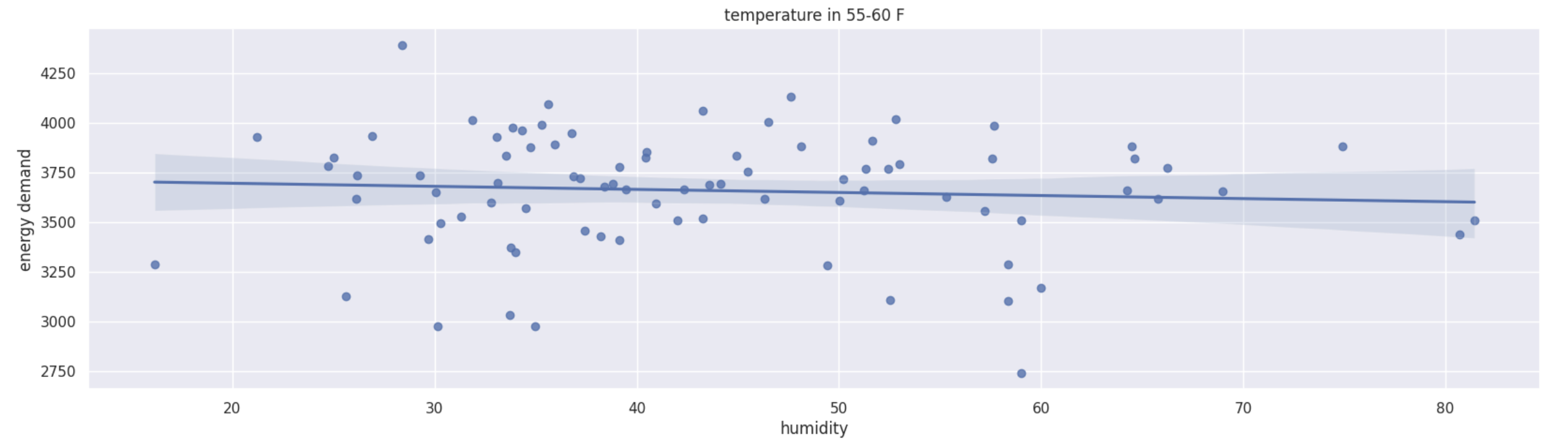}
        \caption{Mild Temperature Scenario}
    \end{subfigure}
    \caption{Humidity vs. Energy Demand: Adjusted for Hour and Month}
    \label{fig:humidity vs energy adjusted}
\end{figure}

In order to address the confounder bias, we need only to control the value of
confounders. Figure \ref{fig:humidity vs energy adjusted} shows the relationship
between humidity and energy demand adjusted for hour and temperature. In the
scenario where temperature is high (above a threshold around 75$^{\circ}$F),
humidity has a positive correlation with energy demand, while this effect
diminishes below the threshold. We remark that the standard deviation of the
energy demand is 369 MW when we condition the temperature to be above
75$^{\circ}$F. Figure \ref{fig:humidity vs energy adjusted}(A) shows that the
effect of humidity can explain a significant portion of the data variability in
this scenario, up to 300 MW if we vary the humidity from the lowest to the
highest. This observation echoes the importance of incorporating humidity in
energy demand forecasting especially in heat events, as pointed out in
\cite{maia2020critical}.

%-------------------------------------
%%%%%%%%%% Move the backdoor path example here
% Another example is temperature vs energy ......(copy paste and revise here)
\subsection{Temperature}\label{subsec:temperature effect}
\begin{figure}[htbp]
    \centering
    \includegraphics[width=1.0\linewidth]{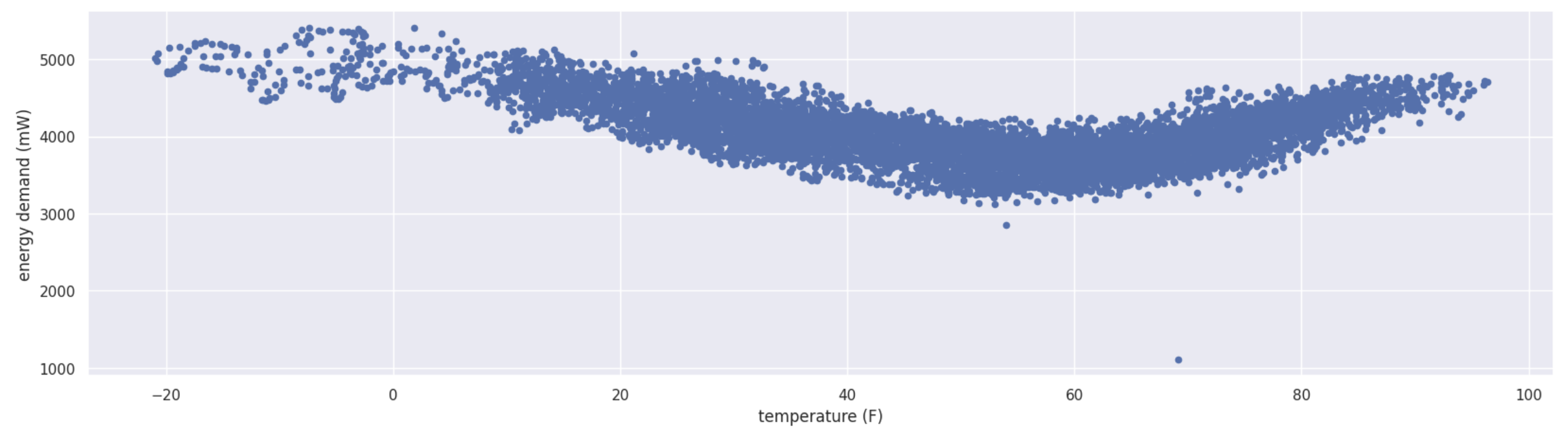}
    \caption{Nonlinear Relation Between Temperature and Energy Demand}
    \label{fig:scatter plot temp vs energy}
\end{figure}

Temperature is known as the most important predictor among weather features. In
various research efforts (e.g., \cite{bessec2008non},
\cite{hekkenberg2009dynamic}), it has been pointed out that energy demand is
more sensitive to temperature changes during summer and winter than it is to
temperature changes during spring and fall. There exists a comfortable
temperature zone (typically around 65-70$^{\circ}$F) away from which the energy
demand increases drastically. As such, spring and fall with milder temperatures
falling inside the comfortable zone are often called ``shoulder seasons" in the
energy industry. In contrast, during summer and winter, when temperatures
deviate substantially from the comfort range, even small temperature changes can
trigger large swings in electricity consumption. This behavior is observed in
our dataset, as shown in Figure \ref{fig:scatter plot temp vs energy}. 

Motivated by Figure \ref{fig:scatter plot temp vs energy}, we transform the raw
temperature using equation \ref{eq: V shape sec 2},
\begin{equation}\label{eq: V shape sec 2}
    \Ttrans = |T - \Tmid|,
\end{equation}
where $\Tmid$ refers to the midpoint between hot and cold extremes when both
heating and cooling needs are minimized (e.g., in the most temperate season for
the geography in question). In this paper, the temperature midpoint $\Tmid$ is
chosen to be $56^{\circ}$F, which best fits the observed data. We further make
the assumption that the energy demand $E$ depends on temperature via
\begin{equation}\label{eq:temp energy model}
    E = \lambda(M) \Ttrans + \text{noise}
\end{equation}
where $M$, ranging from $1$ to $12$, refers to months, and
$\lambda(\text{month})$ is the month-dependent coefficient of the temperature
predictor, reflecting the fact that energy demand responds to temperature
changes differently in each season. To learn these monthly coefficients, we fix
the month and run linear regression based on \eqref{eq:temp energy model}, using
the transformed temperature as the regressor and energy demand as the target.

Note that even after we condition on month, the variable ``hour of day" still
confounds both temperature and energy demand. We have to adjust for its value to
learn the unbiased effect of temperature. This can be accomplished simply by
incorporating hour of day and temperature together in a single regression, as
illustrated in Approach 2 below. In Appendix \ref{sec: appendix}, we show that
this simplified approach is equivalent to conditioning on month of year and hour
of day and estimating the causal effects in the resulting subgroups. This will
be compared with the naive approach as illustrated in Approach 1, where one
attempts to align the transformed temperature and energy demand in a regression
without adjusting for the hour of day as confounder. 

\begin{approach}[Non-causal, learn biased estimate]
\label{approach1}
Perform two regressions. The first linear regression uses $\Ttrans$ as the
regressor and energy demand as the target. A subsequent regression then uses the
trigonometric functions of hours as the regressor, which represent the 24-hour
daily cycle, and the residuals of the first regression as the target.
\end{approach}

\begin{approach}[Causal, no bias]
\label{approach2}
Perform a full regression with temperature and daily cycle as regressors and
energy demand as target. The daily cycle is represented by harmonics
(trigonometric functions) of hours.
\end{approach}

\begin{figure}[htbp]
    \centering
    \includegraphics[width=1.0\linewidth]{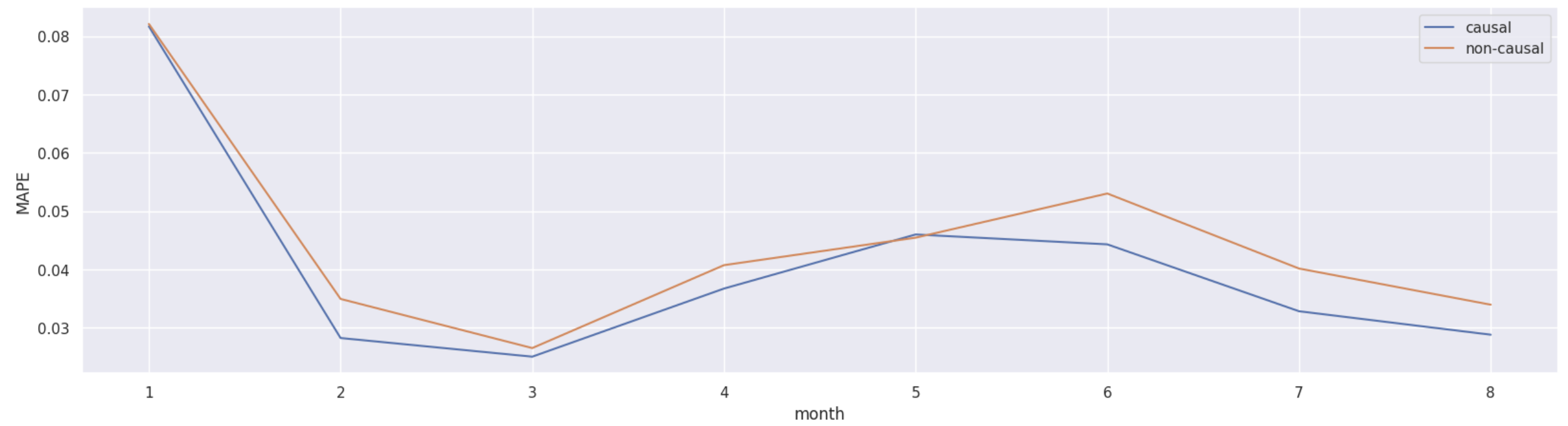}
    \vspace{0.5cm}
    \includegraphics[width=1.0\linewidth]{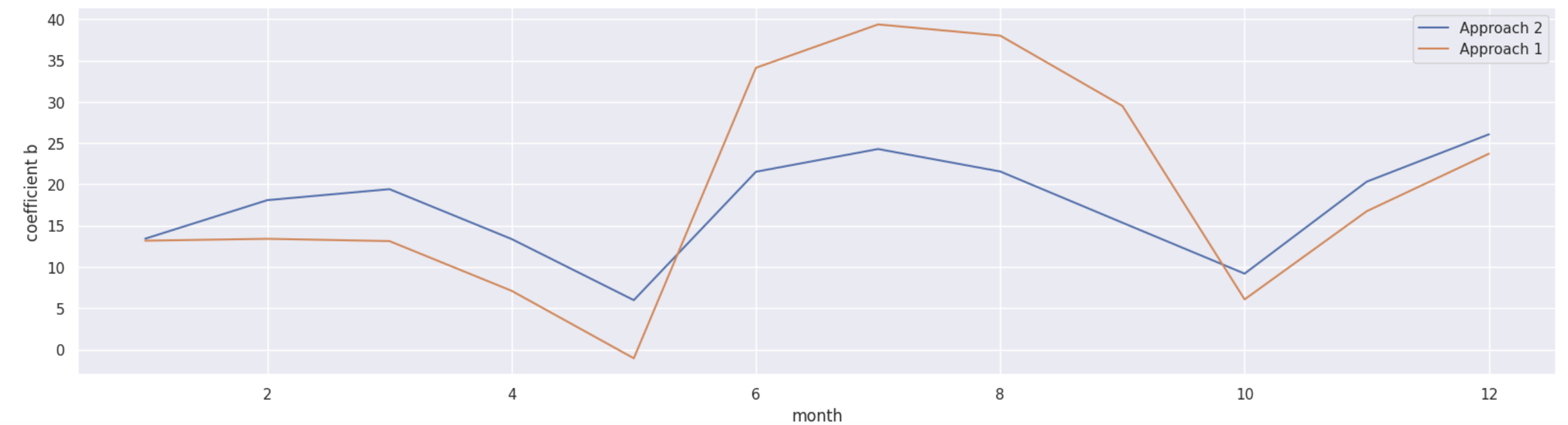}
    \label{fig:causal_vs_non_causal}
    \caption{Estimates of Approach 1 and Approach 2}
    \label{fig:MAPE_causal_vs_non_causal}
\end{figure}

We compare the estimates of these approaches by training both models on 2024
monthly data and testing on 2025 monthly data. The results are evaluated by mean
average percentage error (MAPE), as presented in Figure
\ref{fig:MAPE_causal_vs_non_causal}.

Approach \ref{approach1}, which neglected the causal structure and the existence
of confounding, resulted in 47.8\% deviation in the estimated coefficients
compared to Approach \ref{approach2}, and a 12.5\% worse out-of-sample MAPE on
average. Intuitively, this bias was created due to the facts that
\begin{enumerate}
    \item Temperature affects cooling and heating needs, which drives the energy
    demand directly.
    \item The daily cycle affects the energy demand through two mechanisms:
    \begin{enumerate}
        \item People's work and life schedules depend upon particular times of
        day, which affect energy consumption,
        \item The hour of day affects the temperature, which
        further affects the energy demand.
    \end{enumerate}
\end{enumerate}

\begin{figure}[htbp]
    \centering
    \includegraphics[width=0.4\linewidth]{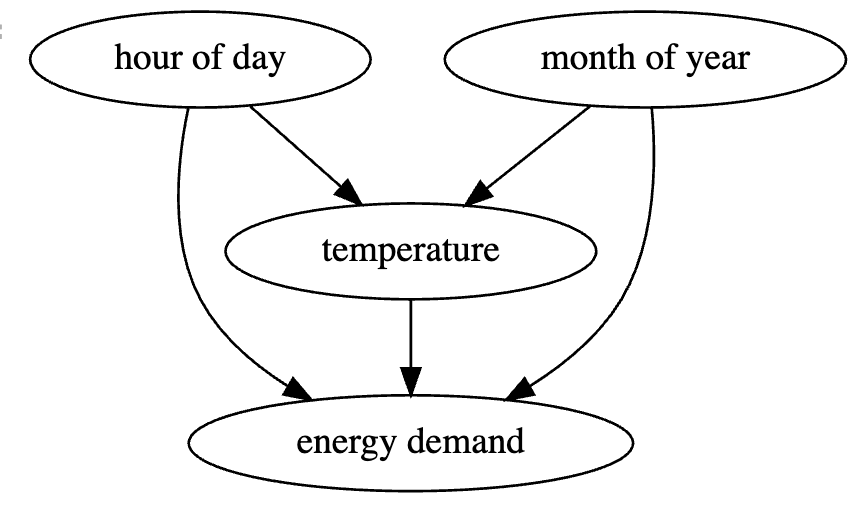}
    \caption{Partial DAG Involving Temperature}
    \label{fig:DAG_1}
\end{figure}

A sample with a high temperature is more likely to be taken in midday, which
happens to be a peak hour in terms of energy consumption by people's daily
activities. On the other hand, samples with low temperature are more likely to
be taken after midnight, when most people are asleep and do not consume energy
due to work/life activities. Without adjusting for hour, Approach
\ref{approach1} incorrectly attributes partial effects of the 24-hour cycle into
that of temperature. In summer, the peak of activity-induced demand aligns with
the peak of cooling demand in midday. This causes Approach \ref{approach1} to
overestimate the effect of temperature in summer. On the contrary, winter
heating demand peaks after sunset when the activity-induced demand declines.
This decoupled effect results in a less volatile energy demand profile in winter
with smaller variance. 

\begin{remark}
    If all of the causal relations in a model are linear, then one can avoid the
    mistake of Approach \ref{approach1} by invoking the Frisch–Waugh–Lovell
    theorem, which states that if the regression we are concerned with is
    expressed in terms of two separate sets of predictor variables:
    \begin{equation}
        Y = X_1 \beta_1 + X_2 \beta_2 + u,
    \end{equation}
    then $\beta_2$ can be learned from a modified regression
    \begin{equation}
        M_{X_1} Y = M_{X_1} X_2 \beta_2 + M_{X_1} u
    \end{equation}
    where $M_{X_1}Y$ is the residual of $Y$ after performing a linear regression
    against $X_1$ and similarly for $M_{X_1}X_2$. Our experiments have shown
    that using the Frisch-Waugh-Lovell theorem yields approximately the same
    results as Approach \ref{approach2}. This can be extended to situations in
    which the target is a linear combination of predictors after feature
    engineering. However, in more general situations where the dependency cannot
    be transformed into linear relations easily, it is essential to recognize
    the underlying causal structure to avoid introducing bias into predictions.
\end{remark}

%-------------------------------------

% 1. hour and month vs temperature 

% 2. temperature vs energy demand (keep the existing causal argument)
% 3. hour and month vs humidity
% 4. humidity vs energy demand (climate specific observations)
% 5. hour and month vs solar radiation
% 6. solar radiation vs energy demand
% 7. wind vs energy demand
\subsection{Solar Radiation}\label{subsec:solar radiation analysis}
\begin{figure}[htbp]
    \centering
    \begin{subfigure}[b]{\textwidth}
        \centering
        \includegraphics[scale=0.3]{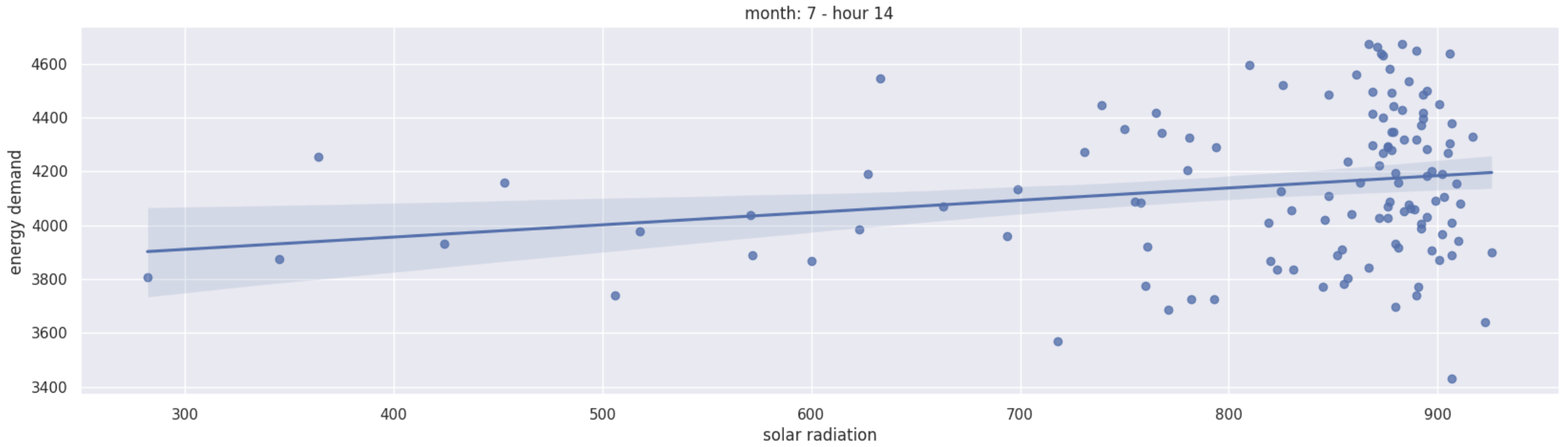}
        \caption{Early Summer Afternoon}
        \label{fig:solar radiation early summer afternoon}
    \end{subfigure}
    \vspace{0.5cm}
    \begin{subfigure}[b]{\textwidth}
        \centering
        \includegraphics[scale=0.3]{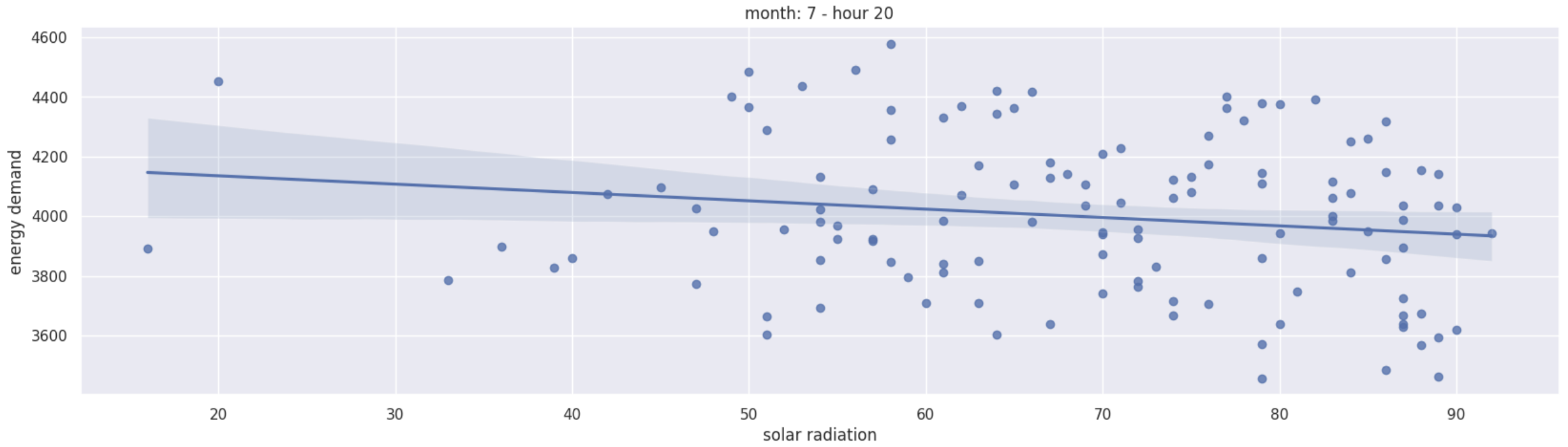}
        \caption{Summer Sunset}
        \label{fig:solar radiation summer sunset}
    \end{subfigure}
    \vspace{0.5cm}
    \begin{subfigure}[b]{\textwidth}
        \centering
        \includegraphics[scale=0.3]{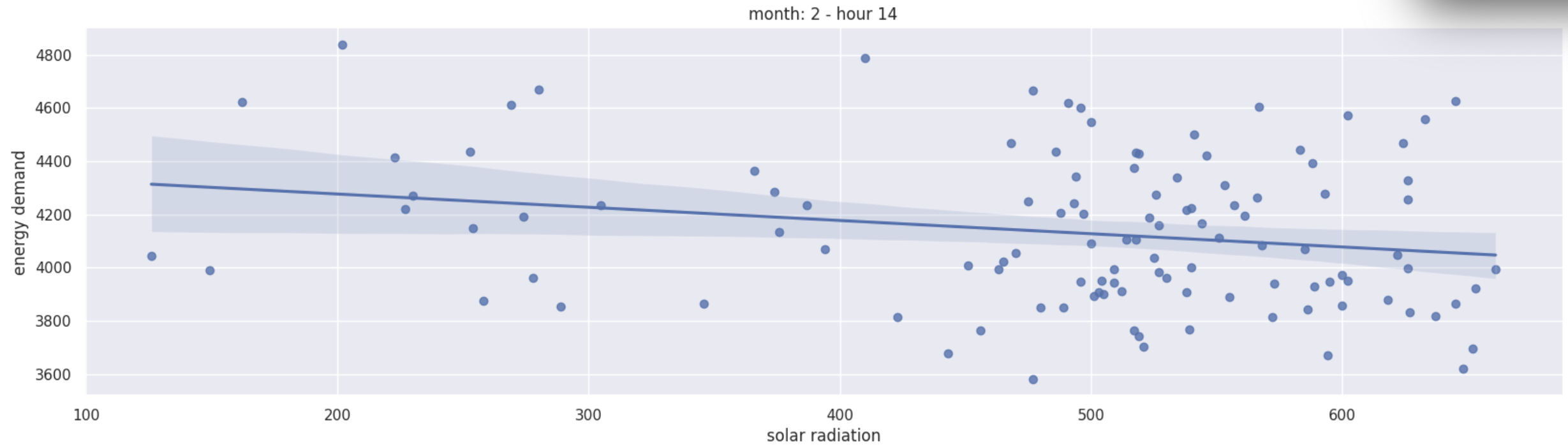}
        \caption{Early Winter Afternoon}
        \label{fig:solar radiation early winter afternoon}
    \end{subfigure}
    \vspace{0.5cm}
    \begin{subfigure}[b]{\textwidth}
        \centering
        \includegraphics[scale=0.3]{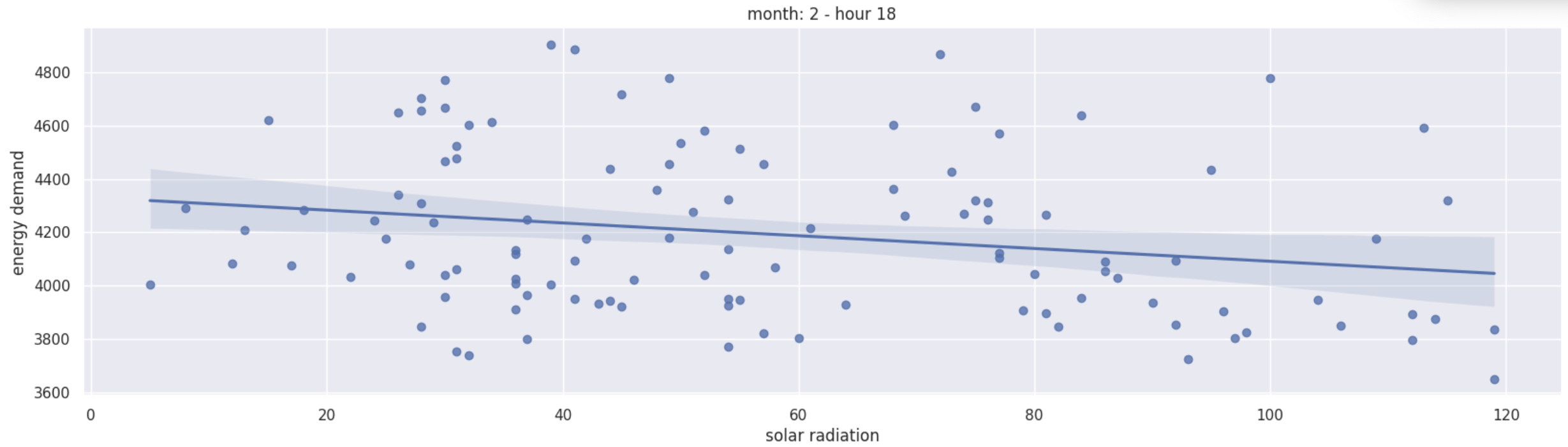}
        \caption{Winter Sunset}
        \label{fig:solar radiation winter sunset}
    \end{subfigure}
    \caption{Solar Radiation vs. Energy Demand}
    \label{fig:solar radiation vs energy demand overall}
\end{figure}

Next, we analyze the effect of solar radiation on energy demand. Intuitively,
insufficient natural light boosts the demand for artificial light. Figure
\ref{fig:solar radiation vs energy demand overall} shows this effect using
samples taken from the month of July and from different hours. It is shown that
the energy demand responds to solar radiation differently in different regimes
similar to the effect of temperature, in the sense that there exists certain
thresholds beyond which further increasing sunlight no longer decreases the
energy consumption. In Figure \ref{fig:solar radiation early summer afternoon},
it even appears as if brighter sunlight increases the energy demand
substantially in summer. This, however, can be easily traced down in the causal
DAG \ref{fig:updated_energy_dag} to the effect of temperature: in the summer,
greater solar radiation warms the air and increases the air temperature, which
translates to greater cooling demand.
As one would expect, during winter, increasing solar radiation leads to warmer
air and thus lower heating needs, as shown in Figure \ref{fig:solar radiation early winter afternoon}.

\subsection{Wind Speed}\label{subsec:wind analysis}
Wind is related to energy demand, as it affects HVAC demand. This effect must be
evaluated jointly with temperature. Figure \ref{fig:wind speed vs energy demand
overall} shows that strong winds decrease energy demand in hot weather and
increase it in cold weather. A possible explanation is that the wind causes
indoor spaces to lose heat faster, which reduce cooling needs in summer and
increase heating needs in winter.

\begin{figure}[htbp]
    \centering
    \begin{subfigure}[b]{\textwidth}
        \centering
        \includegraphics[scale=0.3]{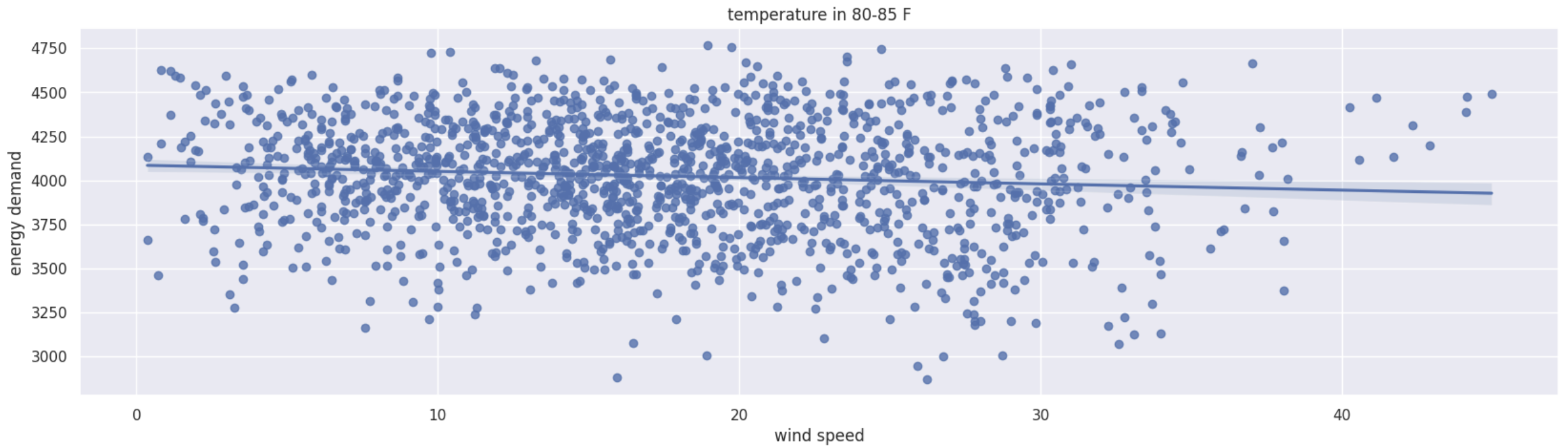}
        \caption{Temperature: $80 \sim 85^\circ{}$F}
        \label{fig:wind speed vs energy summer}
    \end{subfigure}
    \vspace{0.5cm}
    \begin{subfigure}[b]{\textwidth}
        \centering
        \includegraphics[scale=0.3]{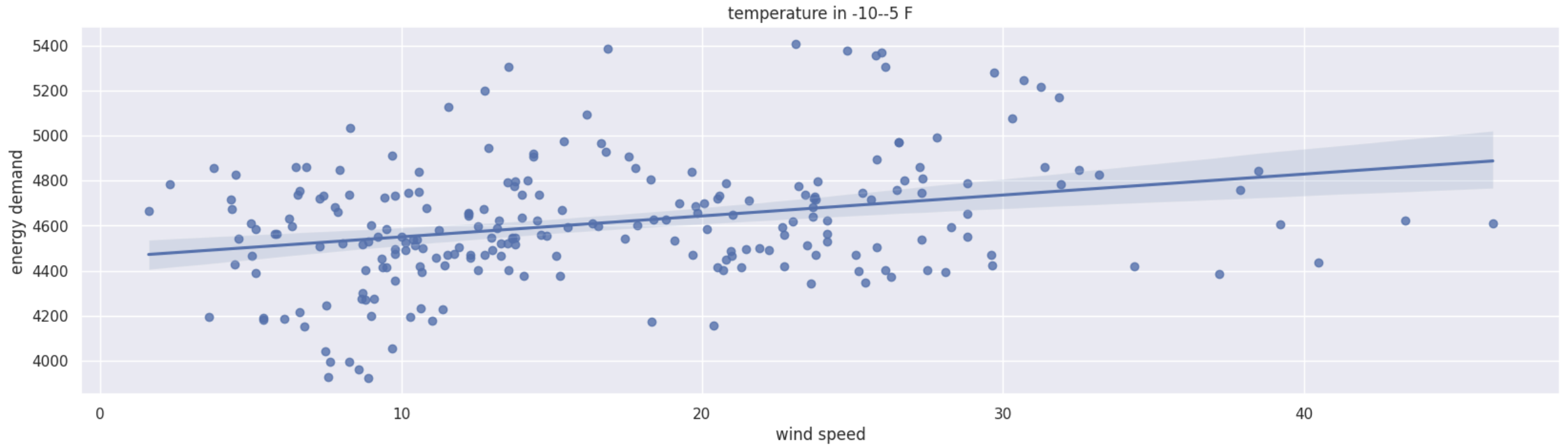}
        \caption{Temperature: $-10 \sim -5^\circ{}$F}
        \label{fig:wind speed vs energy winter}
    \end{subfigure}
    \caption{Wind Speed vs. Energy Demand}
    \label{fig:wind speed vs energy demand overall}
\end{figure}

\section{A Full Bayesian Treatment}\label{sec: bayesian model}
\subsection{Model Setup}\label{subsec:model setup}
We proceed to transform the causal insights derived in Section \ref{sec:causal
vs non causal} into a Bayesian predictive model. Bayesian inference treats model
parameters as random variables with probability distributions rather than fixed
unknown values. One begins with prior beliefs about parameters, then updates
these beliefs using observed data via Bayes' theorem to obtain posterior
distributions that quantify our uncertainty about parameter values. In addition
to providing uncertainty estimates, it can take advantage of prior knowledge,
e.g. derived from domain knowledge or exploratory studies. In our case, we
encode the causal insights from Section \ref{sec:causal vs non causal} into
priors and build a Bayesian model on top.

The model is implemented in Python using the Pyro package
\cite{bingham2019pyro}. It enables us to encode the structural causal model as
represented by Figure \ref{fig:updated_energy_dag} into a DAG. Each node in the
DAG represents an observed variable. The model simulates the data generation
process by a set of manually encoded rules. 
% Describe how the root nodes hour and month were generated.
The hour and month are sampled as categorical variables ranging from 0 to 23 and
1 to 12 respectively. These are root nodes, as they do not have parents in the
DAG. 
% Describe how the childen nodes, e.g. temp, humidity were generated.
The sampling of the child nodes depends upon the value of parent nodes.
Temperature is sampled from a Gaussian distribution whose mean value is a
harmonic series of hour $H$ and month $M$:
\begin{equation}\label{eq:causal temperature}
    \begin{aligned}
        \text{Temperature} & = \overbrace{\sum_{j=1}^{n} (c_j \sin(2\pi j M / 12) + c_j^{*} \cos(2\pi j M / 12))}^{\text{yearly seasons}} \\
        & + \underbrace{\sum_{j=1}^{n} (d_j \sin(2\pi j H / 24) + d_j^{*} \cos(2\pi j H / 24))}_{\text{daily temperature cycle}}
        & + w\cdot \text{Rad}+ T_{\text{base}} + \underbrace{e_{\text{temp}}}_{\text{noise}},
    \end{aligned}
\end{equation}
where $\text{Rad}$ refers to solar radiation, $T_{\text{base}}$ is a constant
baseline and $e_{\text{temp}} \sim \mathcal{N}(0, \sigma^2)$ is Gaussian random
noise.

We sample humidity from a beta distribution $Beta(\alpha, \beta)$ whose
parameters are determined by month and hour.
\begin{equation}
    \begin{aligned}
        & \alpha = \theta \sin(\frac{\pi H}{12}) + \theta^* \cos(\frac{\pi H}{12}) + \alpha_0 \\
        & \beta = \phi \sin(\frac{\pi M}{6}) + \phi^* \cos(\frac{\pi M}{6}) + \beta_0 - \alpha
    \end{aligned}
\end{equation}
The choice of distribution is based on empirical study, which reveals that the
concentration and skewness of the humidity distribution are influenced by month
and hour, see figure \ref{fig:humidity_distribution_vs_hour}, for example, where
the shape of the humidity distribution varies greatly by month and hour.

\begin{figure}[htbp]
    \centering
    \begin{subfigure}[b]{\textwidth}
        \centering
        \includegraphics[scale=0.3]{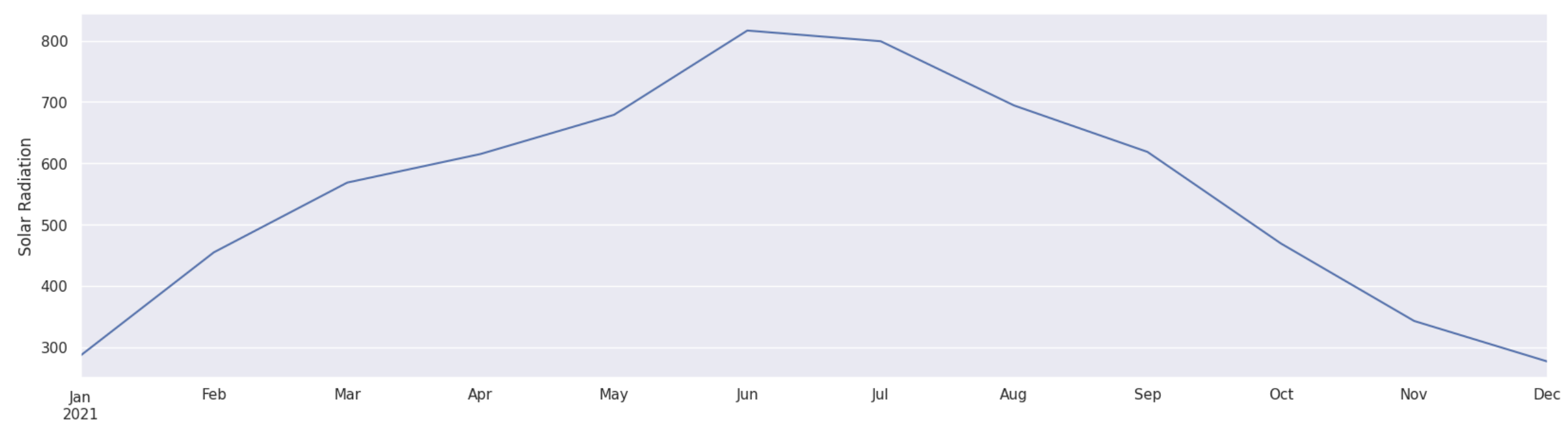}
        \caption{Radiation vs. Month}
        \label{fig:radiation_vs_month}
    \end{subfigure}
    \vspace{0.5cm}
    \begin{subfigure}[b]{\textwidth}
        \centering
        \includegraphics[scale=0.3]{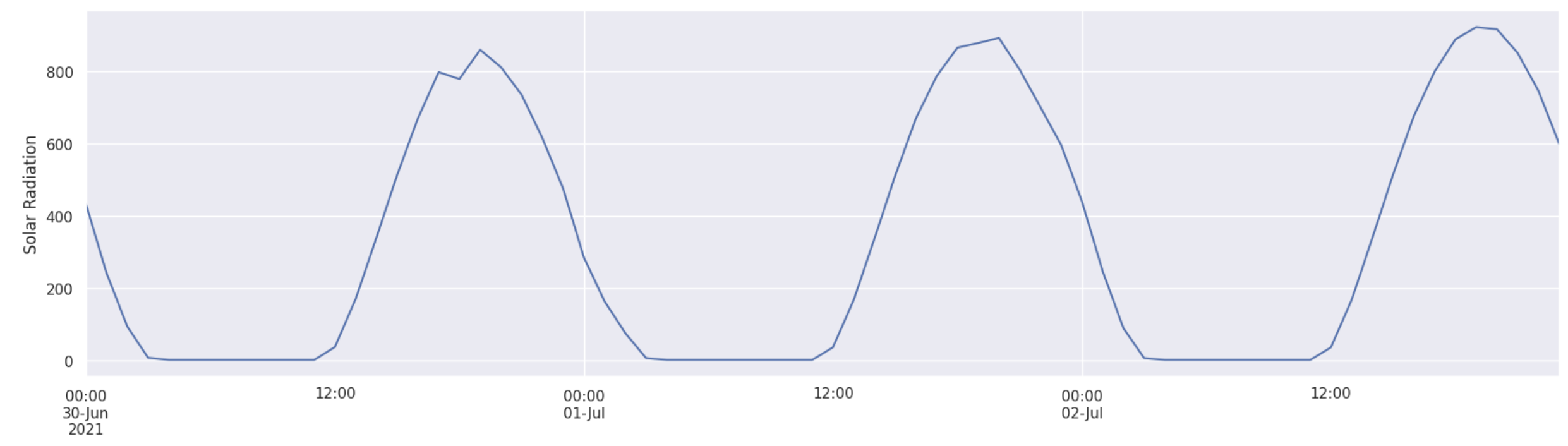}
        \caption{Radiation vs. Hour}
        \label{fig:radiation_vs_hour}
    \end{subfigure}
    \caption{Solar Radiation vs. Calendar Variables}
    \label{fig:radiation_vs_calendar_vars}
\end{figure}

Solar radiation is sampled from a normal distribution and is cast to
non-negative values afterwards. The mean of the normal distribution is defined
as a function of month and hour, where the effects of month and hour are
factorized into separate periodic components.
\begin{equation}\label{eq:radiation_formula}
    \mu_{\text{Rad}} = \left\{ \begin{aligned}
         & (p\sin(\frac{\pi M}{12})+q)\sin(\pi\frac{H-H_{\text{sunrise}}(M)}{H_{\text{sunset}}(M)-H_{\text{sunrise}}(M)}), \\
         & \hspace{1.4cm} if\ H_{\text{sunrise}}(M) \leq H \leq H_{\text{sunset}}(M), \\
         & 0, \hspace{1cm} otherwise.
    \end{aligned} \right.
\end{equation}
where $H_{\text{sunrise}}(M)$ and $H_{\text{sunset}}(M)$ are respectively the
local sunrise and sunset time, averaged over the month $M$. The form of equation
\eqref{eq:radiation_formula} is carefully selected based on the empirical
results, as shown in figure \ref{fig:radiation_vs_calendar_vars}.

Wind speed is sampled directly from normal distribution $\mathcal{N}(\mu_{wind}, \sigma_{wind})$ and then cast to non-negative values. 

% Describe how the energy demand data were generated.
Finally, we define the causal effects of these variables on energy demand. The
proposed causal DAG \ref{fig:updated_energy_dag} states that energy demand can
be divided into three disjoint categories: (1) daily activity consumption, (2)
HVAC needs and (3) lighting needs. We start with HVAC needs, which is mainly
driven by temperature and explains the majority of energy demand variability.
Our analysis in Section \ref{subsec:temperature effect} indicates that the
relationship between temperature $T$ and energy demand can be approximated by a
V-shaped function. In other words, energy demand peaks for both high and low
temperatures, and gradually decreases as temperature moves toward some middle
point. Thus, we define the base energy demand to be
\begin{equation}
    E_{\text{base}} = k f(T) + E_0
\end{equation}
where $f(T) = |T-T_{\text{mid}}|$ is the V-shaped function. The term $E_0$
represents the baseline energy consumption level, which is determined by factors
that stay relatively stable, e.g. population, economy, etc. 

The humidity affects the HVAC needs on top of the base level. In Section
\ref{subsec:humidity}, we learned that relative humidity $R.H.$ increases energy
consumption when the temperature is above some threshold. This effect diminishes
when temperature drops back into the comfort-zone. Thus, we model the effect of
humidity by
\begin{equation}
    E_{\text{humid}} = R.H. \delta(T > T_{R.H.}),
\end{equation}
where $T_{R.H.}$ is the temperature threshold above which the effect of humidity
starts to be visible. In practice, $T_h = 70.0$ is determined via a grid search
by maximizing the correlation between Humidity Induced HVAC and energy demand.

Wind (denoted as $W$) affects energy demand in both high and low temperature
scenarios. As pointed out in Section \ref{subsec:wind analysis}, strong winds
increase energy demand in cold weather and decreases it in hot weathers. We
model the effect of wind by
\begin{equation}
    E_{\text{wind}} = W \delta(T < T_{w,1}) - \lambda W \delta(T > T_{w,2}).
\end{equation}
One unit of change in wind speed may cause different amounts of change in energy
demand depending upon whether the weather is hot or cold. We introduce the
parameter $\lambda > 0$ to model this potential asymmetry. The terms $T_{w,1}$
and $T_{w,2}$ represent the temperature thresholds beyond which the wind's
cooling effects become relevant.

Now we turn to the energy consumption due to daily activities. Since daily
activities are periodic, we model the demand attributable to them using harmonic
series of hours:
\begin{equation}
    \begin{aligned}
        E_{\text{daily}} & = \sum_{j=1}^{n} (a_j \sin(2\pi j H / 24) + a_j^{*} \cos(2\pi j H / 24)).
    \end{aligned}
\end{equation}

Similarly, the effect of the yearly cycle (e.g. including factors like daylight
savings changes) is modeled as a harmonic series of months:
\begin{equation}
    \begin{aligned}
        E_{\text{yearly}} & = \sum_{j=1}^{n} (\alpha_j \sin(2\pi j M / 12) + \alpha_j^{*} \cos(2\pi j M / 12)).
    \end{aligned}
\end{equation}

The lighting demand is expressed as 
\begin{equation}
    E_{\text{light}} = exp(-\beta \cdot \text{Rad}) \delta(\text{active hour}).
\end{equation}
The exponential decay term describes the decreasing need for lighting under
sufficient natural light. The $\delta(\text{active hour})$ declares that the lighting
demand is only present during active hours, e.g. between 5 a.m. and 12 a.m.

Finally, we sum up all of the individual effects:
\begin{equation}
    E = \underbrace{E_{\text{base}} + E_{\text{humid}} + E_{\text{wind}}}_{\text{HVAC needs}} + \underbrace{E_{\text{light}}}_{\text{lighting needs}} + \underbrace{E_{\text{daily}}}_{\text{activity consumption}} + E_{\text{yearly}}
\end{equation}

\subsection{Training}\label{subsec:training}
The specific priors we adopted may be found in Appendix
\ref{appendix:parameters}. After specifying the priors, we condition the
Bayesian model on observed data. This step updates our prior beliefs about
parameters through the likelihood of the observed data and generates posterior
distributions that reflect both our causal assumptions and the observed data.

The posterior distributions are approximated by maximizing the evidence lower
bound (ELBO). The ELBO serves as a surrogate objective that balances the
likelihood of observing the data given a set of parameters against the
complexity of the model. The inference program seeks the set of posterior
parameters that maximizes the ELBO. To perform this optimization, we employ
stochastic variational inference (SVI), a scalable inference method that uses
stochastic gradient descent to approximate the true posterior distribution. An
automatic normal guide (AutoNormal) is used to approximate the posterior
distributions, which assumes a multivariate Gaussian form with a diagonal
covariance structure. The optimization process uses the Adam optimizer with a
learning rate of 0.01.

\subsection{Performance}
The model was trained on data from September 2023 to August 2024 and tested on
data from September 2024 to August 2025. Our model achieves a 3.23\% MAPE on the
training data and 3.84\% MAPE on the test set. 
\begin{figure}[htbp]
    \centering
     \begin{subfigure}{1.0\textwidth} 
        \includegraphics[width=1.0\linewidth]{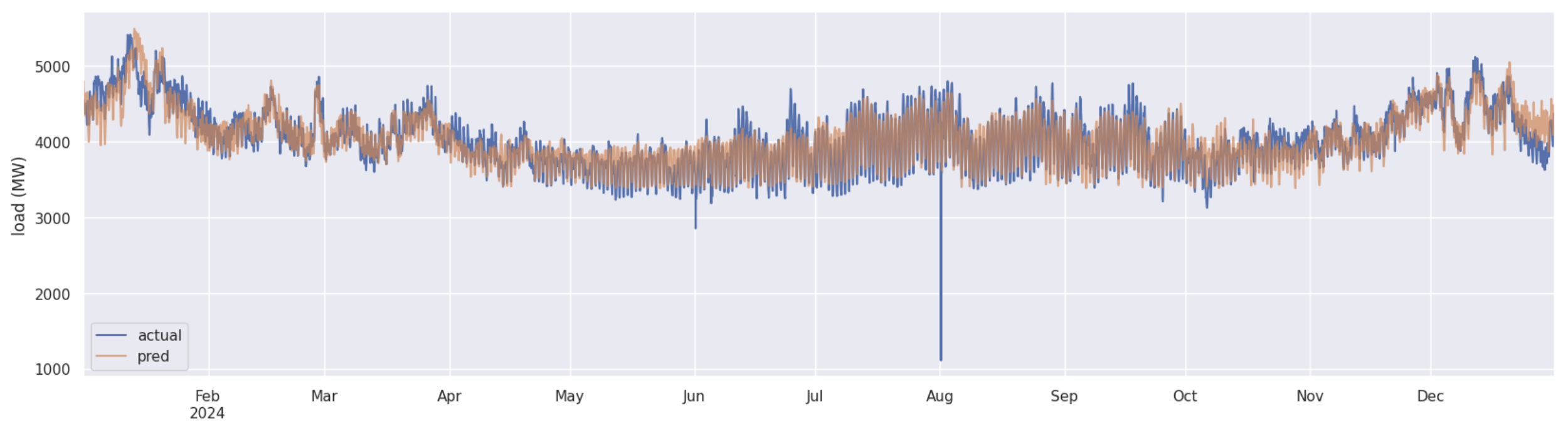}
        \caption{In-sample Test}
    \end{subfigure}
    \vfill
    \begin{subfigure}{1.0\textwidth} 
        \includegraphics[width=1.0\linewidth]{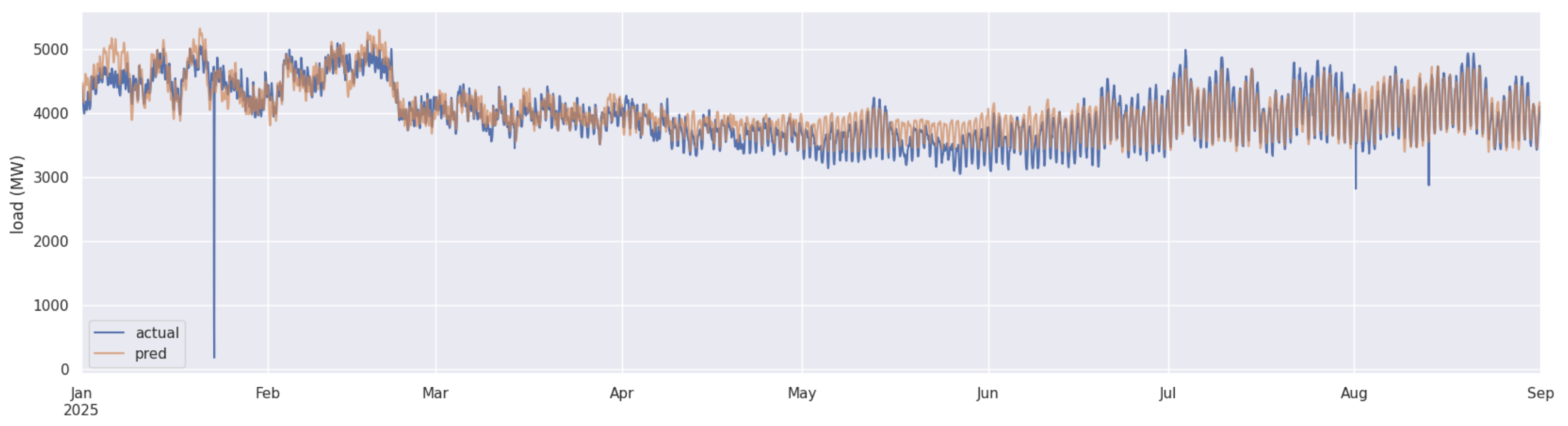}
        \caption{Out-of-sample Test}
    \end{subfigure}
    \caption{Model Performance}
    \label{fig:model performance}
\end{figure}

To validate the robustness of our model's performance, we employed 5-fold
cross-validation, partitioning the union of train and test data into five
distinct folds where each fold serves as a validation set while the remaining
four folds are used for training. This process is repeated five times and the
model's robustness is evaluated by averaging the MAPE across all runs. As a
result, the model yields an average MAPE of 3.88\%, suggesting consistently
strong performance across all folds and no overfitting to specific data
characteristics.

Another notable achievement of our model is that it successfully explains the
time-varying variance (heteroscedasticity) of the energy demand without
requiring explicit variance modeling. We note that electricity demand exhibits
significantly larger variance during summer compared to winter. The mechanism
underlying this phenomenon is the temporal alignment of causal effects as we
pointed out in Section \ref{sec:causal vs non causal}: during summer,
temperature-induced cooling demand and daily-activity-induced energy consumption
both peak around midday, causing these two sources of variation to compound and
amplify aggregate demand volatility. In winter, by contrast, heating demand
peaks during morning and evening hours when temperatures are lowest, while
people's daily activity remains concentrated during midday. This peaking pattern
results in reduced data variance in winter. Compared to pure data driven
approaches, our causal approach provides greater explainability and robustness.

\section{Conclusion}\label{sec:conclusion}
This paper proposed a causal approach for predicting energy demand based on
weather features including temperature, relative humidity, solar radiation and
wind speed, as well as calendar information including hour of day and month of
year. We show that failure to incorporate all the variables that form causal
relations with the target can lead to misleading estimates and attributions. In
some cases (as we have shown in Section \ref{subsec:temperature effect}), not
adjusting for the confounder hour-of-day when estimating the effect of
temperature on energy demand resulted in a staggering 47.8\% bias in the
estimated coefficient and a 12.5\% loss in MAPE. We also presented a full
Bayesian causal model based on the causal structure, which yields robust
performance represented by 3.88\% average MAPE in a K-fold cross validation test
across two years of data.

\section{Future Directions}\label{sec:future directions}
We note that the following directions may further improve the model's
performance.

In one direction, more causal features can be incorporated to explain the data
variability. For example, week of day was not modeled, but could be of
importance for energy demand prediction, since the data pattern is likely to
differ based on whether it is a work day or not.

Incorporating temporal dependencies is another potential direction for
improvement. Electricity consumption is fundamentally a continuous process with
temporal dependencies. When demand exceeds typical levels at time $t$, it tends
to remain elevated at time $t+1$. A natural extension would be to explicitly
model this temporal continuity by incorporating autoregressive (AR) components
or state space models that capture the evolution of demand over time. However,
simply adding lagged demand as a predictor can obscure causal relationships and
potentially introduce confounding. Thus, careful consideration of the causal
structure is required when integrating historical information into causal
models.  

% Namely, we model the causal strength of temperature to energy demand as a sum of harmonics corresponding to month of year seasonality. We fit the updated Bayesian model based on the DAG \ref{fig:DAG_3} on a full year data. Figure \ref{fig:monthly seasonality of b bayesian results} shows the relationship between month of year and the coefficient b.
% \begin{figure}[htbp]
%     \centering
%     \includegraphics[width=1.0\linewidth]{figures/coefficient_with_monthly_seasonality_fit_results.png}
%     \caption{nonlinear relation between temperature and energy demand inferred by Bayesian model}
%     \label{fig:monthly seasonality of b bayesian results}
% \end{figure}

% \section{Performance}
% We may record the performance metrics of different models in this section.

% \section{Direct effects}
% Optionally we can talk about estimations of the amount of direct effect of each cause on the target.

\appendix
\section{Legitimacy of Approach \ref{approach2}}\label{sec: appendix}
In this section, we prove that Approach \ref{approach2} in Section
\ref{subsec:temperature effect} is valid for evaluating causal effects under the
causal assumptions represented by the causal DAG in Figure \ref{fig:DAG_1}.
Specifically, we show that the regression coefficient between temperature and
energy demand derived from Approach \ref{approach2} coincides with the causal
coefficient defined via the intervention operator in \eqref{eq: causal via
intervention}.

\begin{definition}
Causal effects can be quantified via intervention. For a linear model, the
strength of a causal effect can be measured by the coefficient $b$ in
\begin{equation}\label{eq: causal via intervention}
    \E[y | \; do(x)] = bx + c.
\end{equation}
\end{definition}

Let the variables $x, y, z$ denote temperature, energy demand, and hour of day,
just as in Section \ref{sec: bayesian model}. Let $x^*$ be the transformed
temperature. When performing the linear regression using Approach
\ref{approach2}, we essentially assume the following relation: given (condition)
the value of $x$ and $z$, the value of $y$ is a linear combination of $x^*$ and
$y$. No intervention is applied in the process. 
\begin{equation}
    \E[y | x, z] = a^\prime z + b^\prime x^* + c^\prime.
\end{equation}

Note that $b^\prime$ learned from conditioning is not the same as the causal
coefficient $b$ in general. As we have seen in the case of Approach \ref{approach1}, improper
handling of confounding variables will cause bias in our learned results. To
this end, the backdoor criterion provides a means to check if we have handled
the confounders correctly. In fact, the variable $z$ satisfies the backdoor
criterion relative to the ordered pair $(x, y)$ by Definition 3.3.1 in
\cite{pearl2009causality}. Thus we have
\begin{equation}
    P(y | \; do(x)) = \sum_{z} P(y | x, z) P(z).
\end{equation}
Taking the expectation in $z$, we have
\begin{equation}
    \begin{aligned}
        \E[y | \; do(x)] & = \sum_z \E[y | x , z] P(z) \\
        & = \sum_z (a'z + b'x^* + c') P(z) \\
        & = a'\E[z] + b'x^* + c'.
    \end{aligned} 
\end{equation}
This shows that the regression coefficient $b^\prime$ coincides with the
coefficient $b$ in \eqref{eq: causal via intervention}. 

\section{Parameters and Priors}\label{appendix:parameters}
The priors are specified below.

\begin{enumerate}
    \item Parameters for temperature $T$
    \begin{equation*}
        \begin{aligned}
            & c_1 \sim \mathcal{N}(-4.6, \sigma_1) \\
            & c_1^* \sim \mathcal{N}(6.4, \sigma_1) \\
            & c_2 \sim \mathcal{N}(-1.6, \sigma_1) \\
            & c_2^* \sim \mathcal{N}(-0.86, \sigma_1) \\
            & d_1 \sim \mathcal{N}(-17.0, \sigma_3) \\
            & d_1^* \sim \mathcal{N}(-22.0, \sigma_3) \\
            & d_2 \sim \mathcal{N}(-2.3, \sigma_3) \\
            & d_2^* \sim \mathcal{N}(-2.6, \sigma_3) \\
            & w \sim \mathcal{N}(0.01, \sigma_1) \\
            & T_{\text{base}} \sim \mathcal{N}(47.0, \sigma_1) \\
        \end{aligned}
    \end{equation*}
    \item Parameters for humidity:
    \begin{equation*}
        \begin{aligned}
            & \theta \sim \mathcal{N}(0.5, \sigma_1) \\
            & \theta^* \sim \mathcal{N}(-0.7, \sigma_1) \\
            & \phi \sim \mathcal{N}(0.3, \sigma_1) \\
            & \phi^* \sim \mathcal{N}(-0.3, \sigma_1) \\
            & \alpha_0 \sim \mathcal{N}(5.1, \sigma_1) \\
            & \beta_0 \sim \mathcal{N}(7.6, \sigma_1) \\
        \end{aligned}
    \end{equation*}
    \item Parameters for solar radiation:
    \begin{equation*}
        \begin{aligned}
            & p \sim \mathcal{N}(500.0, \sigma_2) \\
            & q \sim \mathcal{N}(300.0, \sigma_2) \\
        \end{aligned}
    \end{equation*}
    \item Parameters for $E_{\text{daily}}$
    \begin{equation*}
        \begin{aligned}
            & a_1 \sim \mathcal{N}(-150.0, \sigma_2^2) \\    
            & a_1^* \sim \mathcal{N}(136.0, \sigma_2^2) \\
            & a_2 \sim \mathcal{N}(84.0, \sigma_2^2) \\
            & a_2^* \sim \mathcal{N}(7.0, \sigma_2^2) \\
        \end{aligned}
    \end{equation*}
    \item Parameters for $E_{\text{yearly}}$
    \begin{equation*}
        \begin{aligned}
            & b_1 \sim \mathcal{N}(-15.0, \sigma_2^2) \\    
            & b_1^* \sim \mathcal{N}(110.0, \sigma_2^2) \\
            & b_2 \sim \mathcal{N}(55.0, \sigma_2^2) \\
            & b_2^* \sim \mathcal{N}(45.0, \sigma_2^2) \\
        \end{aligned}
    \end{equation*}
    \item Parameters for $E_{\text{base}}$
    \begin{equation*}
        \begin{aligned}
            & k \sim \mathcal{N}(20.0, \sigma_2^2) \\
            & E_0 \sim \mathcal{N}(3485.0, \sigma_2^2) \\
        \end{aligned}
    \end{equation*}
    \item Parameters for $E_{\text{humid}}$
    \begin{equation*}
        \begin{aligned}
            & \mu_{\text{humid}} \sim \mathcal{N}(63.0, \sigma_2^2) \\ 
            & \sigma_{\text{humid}} \sim LogNormal(10.0, \sigma_2^2) \\
        \end{aligned}
    \end{equation*}
    \item Parameters for $E_{\text{wind}}$
    \begin{equation*}
        \begin{aligned}
            & \mu_{\text{wind}} \sim \mathcal{N}(16.0, \sigma_2^2),
            & \sigma_{\text{wind}} \sim LogNormal(8.0, \sigma_2^2),
        \end{aligned}
    \end{equation*}
    \item Parameters for $E_{\text{rad}}$
    \begin{equation*}
        \begin{aligned}
            & \mu_{\text{rad}} \sim \mathcal{N}(184.0, \sigma_2^2) \\
            & \sigma_{\text{rad}} \sim LogNormal(167.0, \sigma_2^2) \\
        \end{aligned}
    \end{equation*}
    \item Noise parameters
    \begin{equation*}
        \begin{aligned}
            & \sigma_1 = 4.0 \\
            & \sigma_2 = 40.0 \\
            & \sigma_3 = 10.0 \\
        \end{aligned}
    \end{equation*}
\end{enumerate}

\bibliographystyle{amsplain}
\bibliography{references}

@article{bessec2008non,
  title={The non-linear link between electricity consumption and temperature in {E}urope: {A} threshold panel approach},
  author={Bessec, Marie and Fouquau, Julien},
  journal={Energy Economics},
  volume={30},
  number={5},
  pages={2705--2721},
  year={2008},
  publisher={Elsevier}
}

@book{pearl2009causality,
  title={Causality},
  author={Pearl, Judea},
  year={2009},
  publisher={Cambridge {U}niversity {P}ress},
  edition={2nd},
}

@inproceedings{2018Pearl,
  author={Pearl, Judea},
  title={Theoretical {I}mpediments to {M}achine {L}earning {W}ith {S}even {S}parks from the {C}ausal {R}evolution},
  year={2018},
  isbn={9781450355810},
  publisher={Association for Computing Machinery},
  address={New York, NY, USA},
  url={https://doi.org/10.1145/3159652.3176182},
  doi={10.1145/3159652.3176182},
  booktitle={Proceedings of the {E}leventh {A}{C}{M} {I}nternational {C}onference on {W}eb {S}earch and {D}ata {M}ining},
  pages={3},
  numpages={1},
  location={Marina Del Rey, CA, USA},
  series={WSDM '18},
}

@inbook{2022Schoelkopf,
  author={Sch\"{o}lkopf, Bernhard},
  title={Causality for {M}achine {L}earning},
  year={2022},
  isbn={9781450395861},
  publisher={Association for Computing Machinery},
  address={New York, NY, USA},
  edition={1},
  url={https://doi.org/10.1145/3501714.3501755},
  booktitle={Probabilistic and {C}ausal {I}nference: {T}he {W}orks of {J}udea {P}earl},
  pages={765–804},
  numpages={40},
}

@article{hekkenberg2009dynamic,
  title={Dynamic temperature dependence patterns in future energy demand models in the context of climate change},
  author={Hekkenberg, M and Moll, HC and Uiterkamp, AJM Schoot},
  journal={Energy},
  volume={34},
  number={11},
  pages={1797--1806},
  year={2009},
  publisher={Elsevier}
}

@article{kaddour2022causal,
  title={Causal machine learning: {A} survey and open problems},
  author={Kaddour, Jean and Lynch, Aengus and Liu, Qi and Kusner, Matt J and Silva, Ricardo},
  journal={arXiv preprint arXiv:2206.15475},
  year={2022}
}

@article{yao2021survey,
  title={A survey on causal inference},
  author={Yao, Liuyi and Chu, Zhixuan and Li, Sheng and Li, Yaliang and Gao, Jing and Zhang, Aidong},
  journal={ACM Transactions on Knowledge Discovery from Data (TKDD)},
  volume={15},
  number={5},
  pages={1--46},
  year={2021},
  publisher={ACM New York, NY, USA}
}

@article{jiao2024causal,
  title={Causal inference meets deep learning: A comprehensive survey},
  author={Jiao, Licheng and Wang, Yuhan and Liu, Xu and Li, Lingling and Liu, Fang and Ma, Wenping and Guo, Yuwei and Chen, Puhua and Yang, Shuyuan and Hou, Biao},
  journal={Research},
  volume={7},
  pages={0467},
  year={2024},
  publisher={AAAS}
}

@article{brand2023recent,
  title={Recent developments in causal inference and machine learning},
  author={Brand, Jennie E and Zhou, Xiang and Xie, Yu},
  journal={Annual Review of Sociology},
  volume={49},
  number={1},
  pages={81--110},
  year={2023},
  publisher={Annual Reviews}
}

@article{moraffah2021causal,
  title={Causal inference for time series analysis: {P}roblems, methods and evaluation},
  author={Moraffah, Raha and Sheth, Paras and Karami, Mansooreh and Bhattacharya, Anchit and Wang, Qianru and Tahir, Anique and Raglin, Adrienne and Liu, Huan},
  journal={Knowledge and Information Systems},
  volume={63},
  number={12},
  pages={3041--3085},
  year={2021},
  publisher={Springer}
}

@article{runge2023causal,
  title={Causal inference for time series},
  author={Runge, Jakob and Gerhardus, Andreas and Varando, Gherardo and Eyring, Veronika and Camps-Valls, Gustau},
  journal={Nature Reviews Earth \& Environment},
  volume={4},
  number={7},
  pages={487--505},
  year={2023},
  publisher={Nature Publishing Group UK London}
}

@misc{spp2025hourlyload,
  author = {South West Power Pool},
  title = {Hourly {L}oad {D}ata},
  year = {2025},
  publisher = {South West Power Pool},
  howpublished = {\url{https://portal.spp.org/pages/hourly-load}},
}

@misc{openmeteoapi,
  author = {Open-Meteo},
  title = {Hourly Load Data},
  year = {2025},
  publisher = {Open-Meteo},
  howpublished = {\url{https://open-meteo.com/en/docs/historical-weather-api}},
}

@article{hersbach2020era5,
  title={The {ERA}5 global reanalysis},
  author={Hersbach, Hans and Bell, Bill and Berrisford, Paul and Hirahara, Shoji and Hor{\'a}nyi, Andr{\'a}s and Mu{\~n}oz-Sabater, Joaqu{\'\i}n and Nicolas, Julien and Peubey, Carole and Radu, Raluca and Schepers, Dinand and others},
  journal={Quarterly journal of the royal meteorological society},
  volume={146},
  number={730},
  pages={1999--2049},
  year={2020},
  publisher={Wiley Online Library}
}

@article{bingham2019pyro,
  title={Pyro: {D}eep universal probabilistic programming},
  author={Bingham, Eli and Chen, Jonathan P. and Jankowiak, Martin and Obermeyer, Fritz and Pradhan, Neeraj and Karaletsos, Theofanis and Singh, Rohit and Szerlip, Paul and Horsfall, Paul and Goodman, Noah D.},
  journal={Journal of Machine Learning Research},
  volume={20},
  number={28},
  pages={1--6},
  year={2019}
}

@article{taraldsen2023confidence,
  title={The {C}onfidence {D}ensity for {C}orrelation},
  author={Taraldsen, Gunnar},
  journal={Sankhya A},
  volume={85},
  pages={600--616},
  year={2023},
  publisher={Springer},
  doi={10.1007/s13171-021-00267-y}
}

@article{maia2020critical,
  title={The critical role of humidity in modeling summer electricity demand across the {U}nited {S}tates},
  author={Maia-Silva, Debora and Kumar, Rohini and Nateghi, Roshanak},
  journal={Nature Communications},
  volume={11},
  number={1},
  pages={1686},
  year={2020},
  publisher={Nature Publishing Group UK London}
}
\end{document}